\documentclass[draft]{agujournal2019}
\usepackage{url} 
\usepackage{lineno}
\usepackage{soul}
\usepackage{amssymb}
\usepackage{amsmath}
\usepackage{subcaption}
\usepackage{caption}
\usepackage{booktabs}
\usepackage{dirtytalk}
\justifying

\begin{document}
	
	\title{A Deep-Learning Iterative Stacked Approach for Prediction of Reactive Dissolution in Porous Media}
	
	
	\renewcommand{\thefootnote}{\alph{footnote}}
	\authors{
		Marcos Cirne\affil{1,2}\textsuperscript{\footnotemark}, ORCID: 0000-0001-7266-9211\\
		Hannah P. Menke\affil{1}, ORCID: 0000-0002-1445-6354\\
		Alhasan Abdellatif\affil{1}, ORCID: 0000-0002-4547-378\\
		Julien Maes\affil{1}, ORCID: 0000-0002-3248-9758\\
		Florian Doster\affil{1}, ORCID: 0000-0001-7460-573X\\
		Ahmed H. Elsheikh\affil{1}, ORCID: 0000-0003-3580-7309 
	}
	\footnotetext{Current Affiliation}
	
	\affiliation{1}{School of Energy, Geoscience, Infrastructure and Society (EGIS),\\Heriot-Watt University, EH14 4AS, Edinburgh, United Kingdom}
	
	\affiliation{2}{Biosciences Institute, Newcastle University, NE1 7RU, Newcastle upon Tyne, United Kingdom}
	
	\correspondingauthor{Marcos Cirne}{Marcos.Cirne@ncl.ac.uk}
	\correspondingauthor{Ahmed Elsheikh}{A.Elsheikh@hw.ac.uk}
	

		\begin{abstract}
			Simulating reactive dissolution of solid minerals in porous media has many subsurface applications, including carbon capture and storage (CCS), geothermal systems and oil \& gas recovery. As traditional direct numerical simulators are computationally expensive, it is of paramount importance to develop faster and more efficient alternatives. Deep-learning-based solutions, most of them built upon convolutional neural networks (CNNs), have been recently designed to tackle this problem. However, these solutions were limited to approximating one field over the domain (e.g. velocity field), not accounting for the coupled evolution of multiple interacting fields, including concentration, porosity and flow rates. In this manuscript, we present a novel deep learning approach that incorporates both temporal and spatial information to predict the future states of the dissolution process at a fixed time-step horizon, given a sequence of input states. The overall performance, in terms of speed and prediction accuracy, is demonstrated on a numerical simulation dataset, comparing its prediction results against state-of-the-art approaches, also achieving a speedup around $10^4$ over traditional numerical simulators.
		\end{abstract}
		
		
		%
		%
		
		\section{Introduction}
		\label{sec:intro}
		
		Numerical solvers have been extensively used to simulate and understand the effects of reactive dissolution of solid minerals in subsurface porous media, in diverse applications such as $CO_2$ sequestration \cite{wang2023pore}, hydrogen storage \cite{heinemann2021enabling}, enhanced oil recovery \cite{esfe20203d}, radioactive waste disposal \cite{liang2021review} and geothermal systems \cite{salimzadeh2019coupled}. Due to the intrinsic complexity of these processes, which are governed by a set of highly non-linear partial differential equations (PDEs), it is computationally expensive to simulate \cite{khebzegga2020continuous}. Recently, deep learning (DL) algorithms have become a prominent tool for speeding up the modelling process, while at the same time generating highly accurate simulations of subsurface fluid dynamics \cite{zhu2022review, garnier2021review, da2021deep, kochkov2021machine}.
		
		Typically, DL algorithms for subsurface applications rely on data-driven approaches that require a significant number of examples so that these algorithms can properly learn the underlying physics of the phenomena to be studied. CNN-based methods \cite{alqahtani2018deep, graczyk2020predicting, li2020reaction, santos2020poreflow, tang2021deep}, which employ a series of spatial convolutions to extract meaningful features from image-like data, have been widely adopted to predict specific properties of porous media (such as porosity, permeability and fluid flow). Alternatively, physics-informed neural networks (PINNs) \cite{yan2022physics, du2023modeling, he2020physics} have been adopted to embed the physical laws that govern a given dataset as a prior information into deep neural networks (DNNs).
		
		When it comes to the task of forecasting the future evolution of nonlinear dynamic systems, both categories of neural networks naturally struggle to yield reasonable predictions. Several approaches attempt to combine those types of networks with time-dependent units, including recurrent neural networks (RNNs) \cite{mohajerin2019multistep, ms2021measuring}, convolutional long short-term networks (ConvLSTMs) \cite{cheng2023ensemble, feng2024encoder} and gated recurrent units (GRU) \cite{ding2022low, al2021deep}. A more recent approach named recurrent neural operator (RNO)~\cite{karimi2024learning} was used to map functions rather than discrete data points (thus reducing computational costs) for prediction of reactive flow, but this approach is only applied within large-scale domains. Other have adopted a hybrid approach, such as  \citeA{reichstein2019deep}, who leverages physical process models along with data-driven machine-learning algorithms. The forecasted lead time is bound to the studied phenomenon, and may vary from milliseconds to hundreds of years. However, these methods can only be trained to predict a fixed (limited) amount of future steps, regardless of the time unit.
		
		In order to yield predictions from an initial state for long-term horizons, those dynamic systems employ an iterative strategy in which the output derived from a prediction is used to comprise the input for the subsequent prediction and so on. However, this approach suffers from error propagation with each future step, as the input distributions are more likely to shift away from the distribution under which those systems were trained \cite{koesdwiady2018methods}. Although there are alternatives to avoid the pitfalls of iterative strategies, they are still highly vulnerable to performance degradation due to the intrinsic uncertainties in forecasting further time steps. A possible heuristic to minimize the prediction errors is by stacking multiple networks and perform an iterative process to minimize the overall residual error of the predictions \cite{kani2017dr}. In other words, at each level of the stack, the corresponding network tries to improve the results achieved in the previous level.
		
		This paper presents an approach to predict the dynamical evolution of reactive dissolution in porous media. To the best of our knowledge, this is the first DL-based method in this context that performs simultaneous forecasting of multiple fields. Given an ensemble of numerical pore-scale simulations containing different trajectories for the dissolution process, we train a DL algorithm in a data-driven and supervised way, accounting for spatial and temporal features, as a surrogate model to forecast the future dissolution states. From a sequence of input states, the algorithm is first trained to predict a fixed amount of output states. To assess the quality of the dissolution forecasting, an iterative stacked strategy is adopted and the outputs are evaluated against a ground truth by means of similarity and error metrics. Moreover, a multi-level stacking approach will be investigated as an attempt to reduce the error accumulation. 
		
		The novel contributions of this paper are twofold:
		
		\begin{itemize}
			\item We develop an iterative stacked framework that can be applied to both single-step and multi-step scenarios, while yielding accurate results for reactive dissolution simulations that are orders of magnitude faster than traditional numerical solvers;
			\item We conduct a comparative analysis among different DL architectures in terms of speed and prediction accuracy.
		\end{itemize}
		
		The remainder of this manuscript is organized as follows: Section~\ref{sec:prelim} defines the problem addressed by our work, including the principles of reactive dissolution, forecasting strategies and deep learning algorithms; Section~\ref{sec:method} details our proposed methodology for multi-step prediction of reactive dissolution, as well as the DL methods used in our experiments; Section~\ref{sec:dataset} describes our dataset and the preprocessing steps prior to training the DL algorithms; Section~\ref{sec:results} discusses the produced results; finally, we conclude the paper in Section~\ref{sec:conc}. 
		
		\section{Preliminaries}
		\label{sec:prelim}
		
		\subsection{Problem Statement}
		Let $\mathcal{S} = \{X_1, X_2, ..., X_N\}$, where $X_i \in \mathbb{R}^{C \times H \times W}$, be a numerical simulation of a reactive dissolution process composed of $N$ consecutive time steps of state maps of size $H \times W$ for each of the $C$ physical properties. The governing equations for pore-scale reactive dissolution follow the formulation of \citeA{maes2022improved}. Also, let $m$ be the size of an input sequence of consecutive states and $t$ a starting point from which we want to predict the next $n$ subsequent states. The problem of forecasting consists of taking an input sequence $\mathcal{X} = X_{t-m+1:t} = \{X_{t-m+1}, X_{t-m+2}, ..., X_{t-1}, X_t\}$ and predict an output sequence ${\mathcal{Y}} = \hat{X}_{t+1:t+n}$ based on the input $\mathcal{X}$. In other words, we want to train a deep learning model $\mathcal{F}$ with learnable parameters $\Theta$ which learns a mapping $\mathcal{F}_{\Theta}: \mathcal{X} \mapsto \mathcal{Y}$, where $\mathcal{Y}$ represent the ground truth states, trying to minimize the error between the set of predicted states $\hat{\mathcal{Y}} = \mathcal{F}_{\Theta}(\mathcal{X})$ and the ground truth states $\mathcal{Y}$ according to a loss function $\mathcal{L}$. Therefore, the optimal set of parameters $\Theta^*$ for $\mathcal{F}$ is stated as in Equation~\ref{eq:theta}:
		
		\begin{equation}
			\label{eq:theta}
			\Theta^* = \arg \min_{\Theta}\ \mathcal{L}(\hat{\mathcal{Y}}, \mathcal{Y})
		\end{equation}
		
		We refer as \textbf{multi-step forecasting} the cases where $n > 1$, and as \textbf{single-step forecasting} when $n = 1$.
		
		\subsection{Forecasting Strategies}
		An in-depth analysis of possible forecasting strategies for multivariate time-series data can be found in \citeA{lim2021time}. Among the existing strategies, there is \textbf{direct forecasting}, in which a DL model is trained to forecast each of the $n$ future steps~\cite{makridakis2018statistical}, but it does not consider the relationships among the predicted states $\hat{X}_{t+i}$, also being limited to a maximum forecast horizon $n$. Another strategy is known as \textbf{iterative forecasting} (also called \textbf{recursive forecasting}), in which a trained model outputs one step ahead, then uses this output to comprise the inputs for the prediction of the next step, which yields the output for the subsequent step, and so on. However, this strategy is highly prone to error accumulation, especially for longer horizons in which all inputs are forecasted values rather than actual observations \cite{taieb2012review}.
		
		To circumvent those issues, there are alternate strategies that predict multiple steps at the same time. The most widely used is known as \textbf{Multiple-Input Multiple-Output (MIMO)}, which preserves the dependency between the forecasted values, also reducing the error accumulation problem from the iterative strategy (up to time step $t+n$). 
		
		
		
		Besides being applied in time-series forecasting, this strategy has been widely used for spatiotemporal forecasting tasks, such as video prediction \cite{gao2022simvp,oprea2020review} and earth science forecasting \cite{xu2021spatiotemporal,nguyen2023scaling}. Even though the forecasting horizon is limited by the model, one can perform recursive steps to yield predictions for longer time steps.
		
		
		
		
		
		\subsection{Deep Learning Methods}
		\label{subsec:dl}
		
		\subsubsection{Encoder-Decoder ConvLSTMs}
		
		Convolutional Long Short-Term Memory Networks (ConvLSTMs) \cite{shi2015convolutional} are a class of neural networks designed to capture spatiotemporal dependencies in a time-evolving sequence, combining the strengths of CNNs and LSTMs into a single method. Given an input sequence $\mathcal{X}_t$, as well as the hidden state $\mathcal{H}_{t-1}$ and the cell state $\mathcal{C}_{t-1}$ from the previous time step, the gate mechanisms contained in a ConvLSTM cell -- represented by $i_t$ (input gate), $f_t$ (forget gate), and $o_t$ (output gate) -- control the amount of information that is going to be propagated (or forgotten) for the next states. The process of computing the next hidden and cell states from an input sequence $\mathcal{X}$ can be mathematically described as shown in Equation~\ref{eq:convlstm}, where $\sigma$ is the sigmoid activation function, $*$ represents the convolution operations, $\circ$ stands for the Hadamard product (element-wise product), $W$ and $b$ are weights and biases for each of the gates and the cell state.
		
		\begin{align}
			\label{eq:convlstm}
			i_t &= \sigma\left(W_{xi} * \mathcal{X}_t + W_{hi} * \mathcal{H}_{t-1} + W_{ci} \circ \mathcal{C}_{t-1} + b_i\right) \nonumber \\
			f_t &= \sigma\left(W_{xf} * \mathcal{X}_t + W_{hf} * \mathcal{H}_{t-1} + W_{cf} \circ \mathcal{C}_{t-1} + b_f\right) \nonumber \\
			\mathcal{C}_t &= f_t \circ \mathcal{C}_{t-1} + i_t \circ \tanh\left(W_{xc} * \mathcal{X}_t + W_{hc} * \mathcal{H}_{t-1} + b_c\right) \\
			o_t &= \sigma\left(W_{xo} * \mathcal{X}_t + W_{ho} * \mathcal{H}_{t-1} + W_{co} \circ \mathcal{C}_t + b_o\right) \nonumber \\
			\mathcal{H}_t &= o_t \circ \tanh(\mathcal{C}_t) \nonumber
		\end{align}
		
\begin{figure}[h]
			\centering
			\includegraphics[width=1\linewidth]{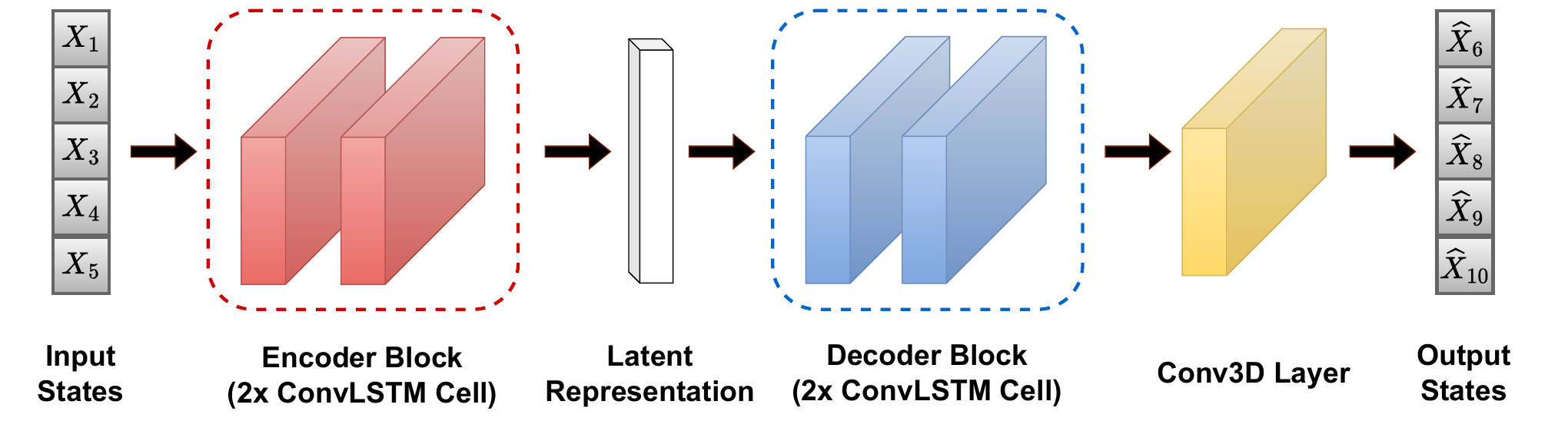}
			\caption{Encoder-Decoder structure of ConvLSTM cells for multi-step prediction.}
			\label{fig:convlstm}
		\end{figure}

		To perform multi-step prediction with ConvLSTMs, a sequence of ConvLSTM cells is structured as an encoder-decoder architecture (also known as \textbf{Seq2Seq} model), where both encoder and decoder blocks contain the same number of ConvLSTM cells. This type of architecture is inspired on sequence-to-sequence (Seq2Seq) models for natural language processing and time-series tasks \cite{bahdanau2014neural, sutskever2014sequence}. An example is depicted in Figure~\ref{fig:convlstm}, with two ConvLSTM cells in each block. After an input sequence $\mathcal{X}$ is processed by the encoder block, the final hidden state from the last ConvLSTM cell forms a compact representation of $\mathcal{X}$ (latent representation), which is fed to the decoder block to produce the hidden states for each output time step. Finally, a 3D convolutional layer receives the output from the decoder block and produces the predictions of all physical properties for each desired time step. More details about the encoding and decoding processes can be found in \cite{kakka2022sequence}.
		
		\subsubsection{U-Shaped Fourier Neural Operator}
		Proposed by \citeA{wen2022u}, the U-Shaped Fourier Neural Operator (U-FNO) is an extension of the Fourier neural operator (FNO) \cite{li2020fourier}, designed to solve PDEs across diverse problems involving computational fluid dynamics. FNOs are known to be resolution-invariant, meaning that they can be trained on a lower resolution and evaluated on higher resolution, and yield superior performance against conventional CNNs by operating directly on the Fourier space (frequency domain), replacing convolution operations by pointwise multiplications, which are much faster and efficient.
		
		To increase the overall performance in multiphase flow problems, the aforementioned authors introduced the U-Fourier layer as an upgrade of the original FNO architecture, combining the advantages of both CNN- and FNO-based models, increasing both training and test accuracies. On the other hand, this improvement on the accuracies comes in expense of the flexibility of training and testing at different resolutions. Moreover, U-FNO's usually take a longer time to be trained than traditional FNO's. 
		
		
		\subsubsection{Temporal Attention Unit}
		
		Developed for video prediction tasks, Temporal Attention Units (TAUs)~\cite{tan2023temporal} leverage the ability to capture time evolution in image sequences by introducing parallelizable attention mechanisms that eliminate the need of recurrent-based units (such as RNNs and ConvLSTMs), speeding up the training process. In turn, the spatial modules are represented by simple 2D convolutions. Moreover, not only TAUs account for intra-frame differences through the mean squared error loss, but also for inter-frame variations by embedding a differential divergence regularization term. The resulting loss function is expressed in Equation~\ref{eq:tau}:
		
		\begin{equation}
			\label{eq:tau}
			\mathcal{L}(\hat{\mathcal{Y}}, \mathcal{Y}) = \sum ||\hat{\mathcal{Y}} - \mathcal{Y}||^2 + \alpha \mathcal{L}_{reg}(\hat{\mathcal{Y}}, \mathcal{Y})
		\end{equation}
		
		\noindent where $\mathcal{L}_{reg}$ represents the Kullback-Leibler divergence between the probability distributions of the inter-frame differences from $\hat{\mathcal{Y}}$ and $\mathcal{Y}$, and $\alpha$ is a weight term defined empirically.
		
		
		\section{Methodology}
		\label{sec:method}
		
\begin{figure}[h]
			\centering
			\includegraphics[width=0.95\linewidth]{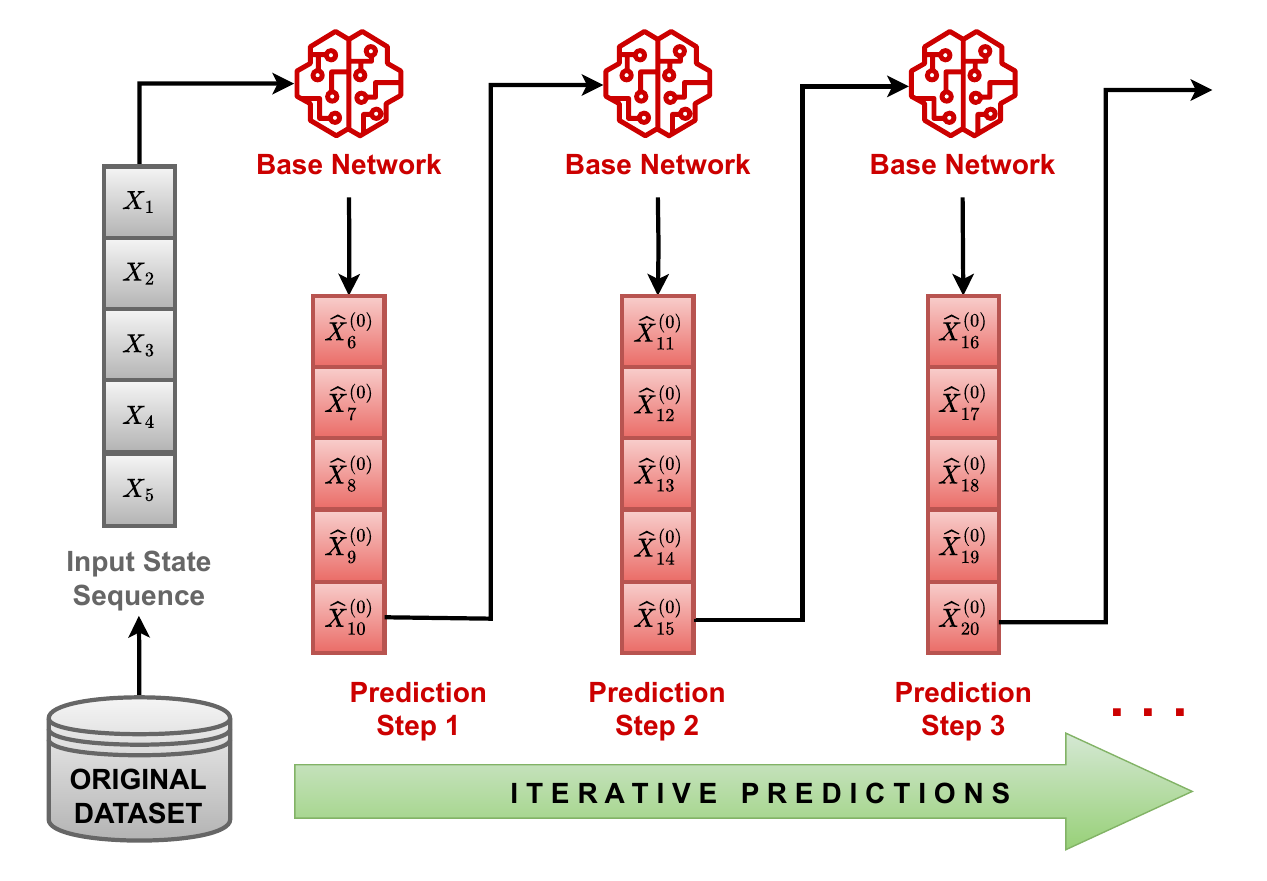}
			\caption{Flowchart of our iterative multi-step approach with network stacking for full evaluation of a reactive dissolution process on simulation data.}
			\label{fig:method_1}
		\end{figure}

		Assuming $m = 5$ input steps and $n = 5$ output steps, we first train a base network by making it receive a sequence of 5 \say{perfect} input states and predict the subsequent 5 states (i.e., without successive iterations), as done in a traditional MIMO approach. After the training process, to evaluate the full evolution of the dissolution process on a given simulation $\mathcal{S}$ containing $N$ total steps, we conduct successive iterative predictions, as illustrated in Figure~\ref{fig:method_1}, by taking the first 5 input states from $\mathcal{S}$, yielding outputs at time steps 6-10. In turn, this output is fed as an input to the same base network to produce the states at time steps 11-15, and so on. 
		
		To improve the initial solution produced by the base network, we propose a multi-level stacking of trained networks. In this case, the same neural network architecture is used to train each level, with the same sizes for both input and output tensors as the base network, as well as the loss function. Considering the base network as Level 0, trained with a dataset $\mathcal{D} = \{\mathcal{S}_1, \mathcal{S}_2, ..., \mathcal{S}_{ts}\}$, where $ts$ is the number of training samples, we generate a new dataset comprised of all possible output sequences $\mathcal{\hat{Y}}$ computed by the base network for each $\mathcal{S}_i \in \mathcal{D}$, representing initial approximations to the true states $\mathcal{Y}$. The Level 1 network is then trained to receive, as input, one of those responses produced by the base network, and outputs a new approximation (correction) to the ground truth within the same time-step interval, hypothesizing that it will yield less errors than the initial solution. Following this rationale, we can continue the stacking process by training a Level 2 network with the outputs produced by Level 1, and so on until no further improvement is observed. During this process, only one level can be trained at a time.
		
\begin{figure}[h]
			\centering
			\includegraphics[width=\linewidth]{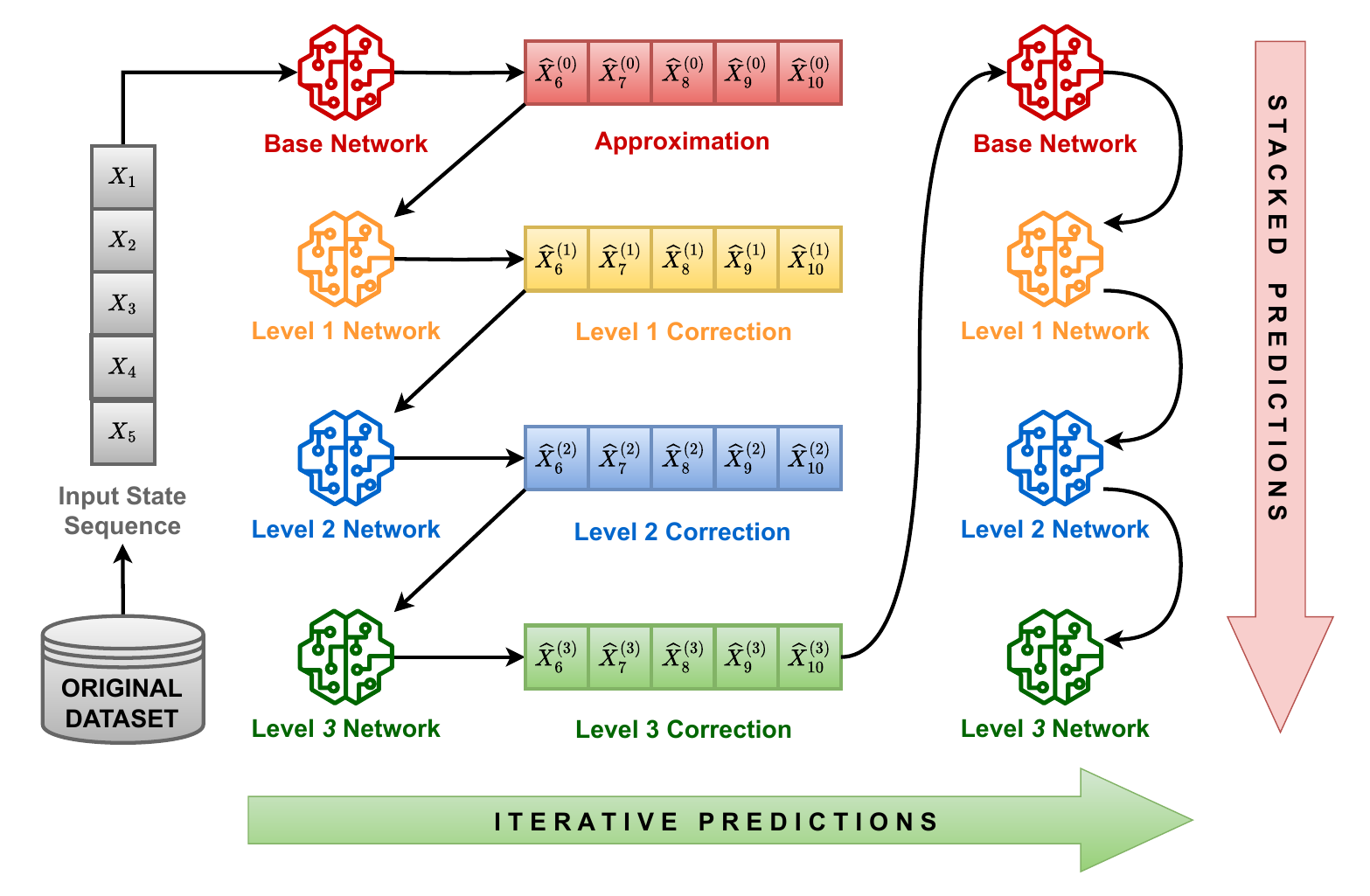}
			\caption{Flowchart of our iterative multi-step approach with network stacking for predicting reactive dissolution.}
			\label{fig:method_2}
		\end{figure}

		In general, let $L$ be the number of levels of correction to be applied on the base solution. After the base and all of the $L$ correction networks are trained, the iterative multi-step prediction process can be conducted as illustrated in Figure~\ref{fig:method_2} with $L = 3$. Here, the first 5 input states of a simulation sample are fed into the base network, producing an initial approximation for the subsequent 5 states. Then, this output is fed to the Level 1 network which produces the first level of correction for the initial approximation. Later, this output serves as input to Level 2 network, whose output is forwarded as an input to Level 3 network, which yields the final prediction for time steps 6-10. For the next prediction step, the output from the last network of the stack is fed to the base network to produce an initial approximation for steps 11-15, which is corrected by the subsequent levels in the stack. The whole process is repeated until the simulation end time.
		
		\section{Reactive Dissolution Dataset and Preprocessing}
		\label{sec:dataset}

		Our reactive dissolution dataset~\cite{reactdataset2025} consists of 32 numerical simulations generated using \texttt{GeoChemFoam}~\cite{maes2022improved}, an open-source micro-continuum solver for pore-scale reactive transport. Each simulation follows the \textit{Benchmark~2} configuration presented by~\citeA{maes2022improved}, which models calcite dissolution in a quasi two-dimensional micromodel. The computational domain consists of a polydisperse disk pack with an initial porosity of approximately $\phi_0 = 0.45$ and an initial permeability of $K_0 = 5.6\times10^{-10}\,\mathrm{m^2}$. Simulations are performed on a uniform Cartesian mesh with a spatial resolution of $\Delta x,y = 25\,\mu\mathrm{m}$ and a $\Delta z = 150\,\mu\mathrm{m}$ . 
		
		The flow field is incompressible and governed by the Darcy-Brinkman-Stokes equations under isothermal conditions, while the transport of a single reactive species obeys an advection--diffusion equation with a surface reaction rate $R = k_c\,c$ applied at the fluid--solid interface. A representative parameter set corresponding to the \textit{conical wormhole regime} is as follows: kinematic viscosity $\nu = 1\times10^{-6}\,\mathrm{m^2/s}$, diffusion coefficient $D = 1\times10^{-9}\,\mathrm{m^2/s}$, inlet concentration $c_i = 0.01\,\mathrm{kmol/m^3}$, stoichiometric coefficient $\zeta = 1$, calcite density $\rho_s = 2710\,\mathrm{kg/m^3}$, molar mass $M_{ws} = 100\,\mathrm{kg/kmol}$, Kozeny--Carman constant $k_f = 2\times10^{11}\,\mathrm{m^2}$, a volumetric flow rate $Q=5.2174 \times10^{-12}\,\mathrm{m^3/s}$, and  reaction rate constant $k_c = 8.1632\times10^{-6}\,\mathrm{m/s}$. These conditions correspond to Péclet and kinetic numbers of $\mathrm{Pe}=1$ and $\mathrm{Ki}=1$, respectively, representing the transition between compact and wormhole dissolution regimes. Boundary conditions follow~\citeA{maes2022improved}: a constant inflow rate at the inlet, fixed outlet pressure, and no-slip/no-flow boundaries on the top and bottom walls.
		
\begin{figure}[h]
			\centering
			\includegraphics[width=\linewidth]{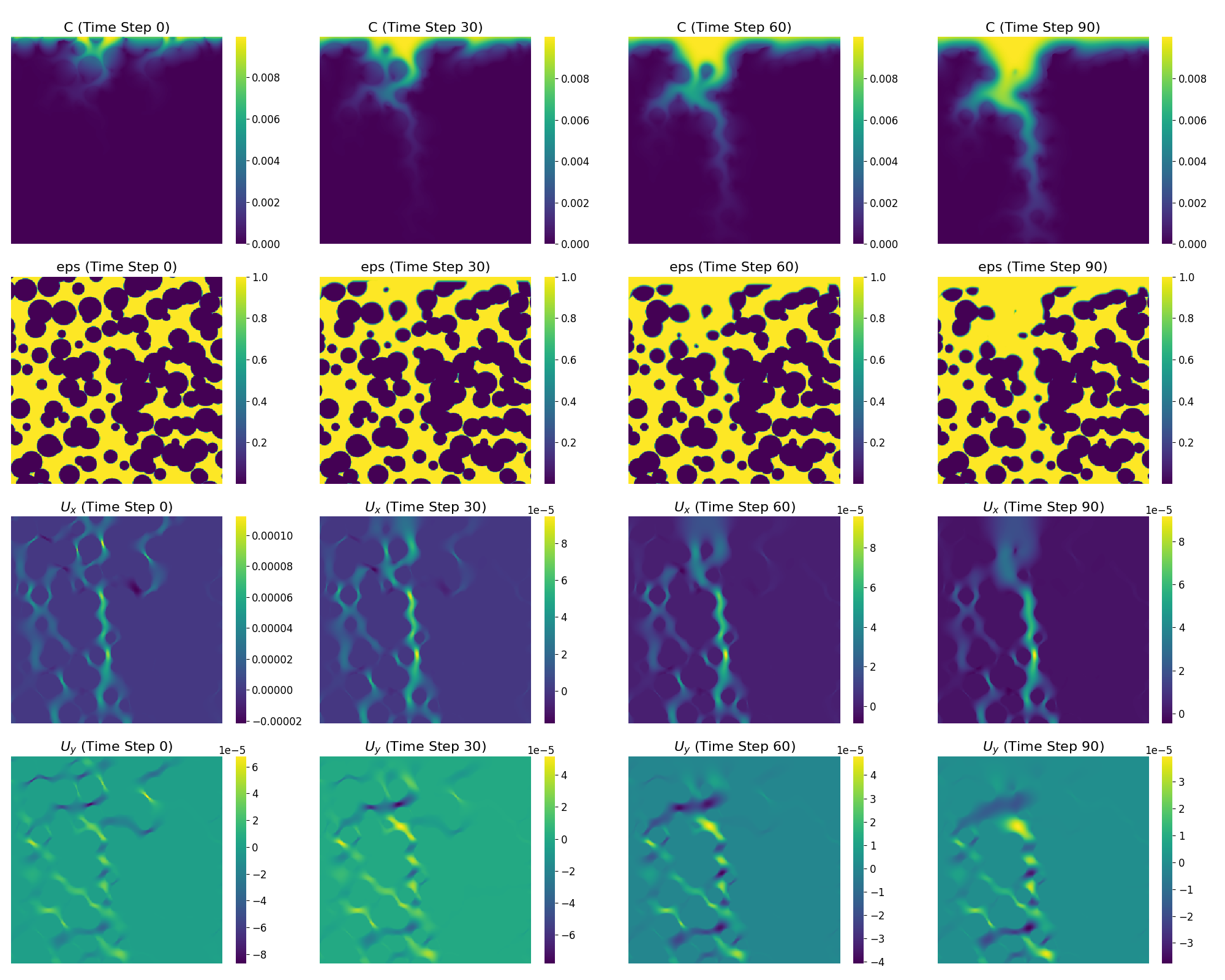}
			\caption{Input features from one dataset sample representing the dissolution states for time steps 0, 30, 60 and 90 (from left to right).}
			\label{fig:inp_features}
		\end{figure}

		The 2D simulation results are stored at 100 equally spaced time steps, resulting in time-evolving dissolution state maps of size $260 \times 260$. For the ML algorithms tested in this work, all maps were cropped to $256 \times 256$ by removing the first and last two rows and columns from each state map. Each of these simulations conveys a particular rock sample with its own pore structure, as well as a distinct fluid trajectory over the course of the dissolution process. An example is illustrated in Figure~\ref{fig:inp_features}. These maps encompass four different input properties: $\mathbf{C}$, the concentration of the acidic solution used in the dissolution; $\mathbf{eps}$, a function indicating the volume fraction of the pore space occupied by pore in each voxel, which is proportional to the amount of fluid contained in each voxel; $\mathbf{U}_x$, the magnitude and direction of flow of the acidic solution in the horizontal axis of a 2-D Cartesian plane; and similarly $\mathbf{U}_y$, for the vertical axis. The average time to produce each simulation was approximately 3 hours using 24 CPUs of 3 GHz each. All simulations employ the same flow and reactive transport conditions, where both P\'{e}clet and kinetic numbers are equal to 1.
		
		In essence, GeoChemFoam solves the reactive transport in porous media in a quasi-static regime, as described by~\citeA{maes2022improved}. Given a porosity map $\mathbf{eps}$ at a time step $t$, it first calculates the steady-state velocity (i.e., it produces $\mathbf{U}_x$ and $\mathbf{U}_y$ at time step $t$). Then, it calculates a steady-state concentration $\mathbf{C}$ at time step $t$, which includes the contributions of injection, diffusion, and reactive dissolution at the fluid-solid interface. From these calculations, it estimates a local reaction rate in each computational fluid and the new $\mathbf{eps}$ field for the time step $t+1$, which in turn is used to calculate the flow once again to produce a new $\mathbf{eps}$ map at time step $t+2$, and so on.
		
		Aside from the original features, three extra input features were engineered as an attempt to improve the performance of all DL models: 1) \textbf{Magnitude of velocity}: defined as $U = \sqrt{U_x^2 + U_y^2}$; 2) {$\mathbf{C}$} \textbf{Scaled}, a log-transformation over the concentration values; and 3) \textbf{Combined Filter}, a binary mask based on $C$ and $eps$ constraints which shows the portions of the grains in a porous matrix that are being dissolved at a particular time step $t$. Given a position $(i,j)$ in a map, the combined filter is calculated according to Equation~\ref{eq:filter}:
		
		\begin{equation}
			\text{Filter}(i,j) = 
			\begin{cases} 
				1 & \text{if } C(i, j) \geq 10^{-4} \text{ and } 0.01 \leq eps(i,j) \leq 0.99, \\
				0 & \text{otherwise}.
			\end{cases}
			\label{eq:filter}
		\end{equation}
		
		\section{Performance Evaluation of Deep Learning Models}
		\label{sec:results}
		
		In this section, we will discuss the quantitative and qualitative results of our proposed method on all algorithms described in Section~\ref{subsec:dl}. The source code is available at \url{https://github.com/ai4netzero/ReactiveDissolution}.
		
		\subsection{Training and Evaluation Settings}
		From the 32 simulation samples in our dataset, 24 were randomly selected as the training set and the remaining 8 as the validation/test set. Before the DL models are trained, the entire data is normalized with respect to the mean and standard deviation values of the training set. All input features, except for C Scaled and the Combined Filter, have their original values subtracted by their respective means and the results are each divided by their respective standard deviations. On the other hand, a min-max normalization is applied to the output values so that they all belong within the $[0, 1]$ range, making it suitable to adopt the sigmoid activation function for all DL models.
		
		Concerning our proposed multi-level stacking of neural networks method, for each DL algorithm described in Section~\ref{subsec:dl}, we perform corrections up to Level $L = 3$. The models were trained on a NVIDIA Titan RTX with 24 GB of memory for a total of 100 epochs (along with a patience rate of 20 epochs for early stopping), using a batch size of 4, a learning rate of 0.0005 and the Adam optimizer \cite{kingma2014adam} with moment values $\beta_1 = 0.9$ and $\beta_2 = 0.999$. The mean squared error was adopted as the loss function for ConvLSTM, U-FNO and the intra-frame difference term of TAU. Regarding the latter, the $\alpha$ constant for the regularization term (inter-frame difference) was fixed at 0.1.
		
		For the quantitative evaluation of the iterative predictions of each output property, we calculate the Pearson correlation coefficient (PCC) to assess the similarity between the predicted state $\hat{Y} \in \mathbb{R}^{H \times W}$ and the ground truth state $Y \in \mathbb{R}^{H \times W}$ at each time step, as stated in Equation~\ref{eq:pcc}:
		
		\begin{equation}
			\label{eq:pcc}
			\text{PCC}({\hat{Y}, Y}) = \frac{\text{cov}(\hat{Y}, Y)}{\sigma_{\hat{Y}} \cdot \sigma_{Y}}
		\end{equation}
		
		\noindent where cov$(\hat{Y}, Y)$ is the covariance between $\hat{Y}$ and $Y$ and $\sigma$ represents the standard deviation of a state map.
		
		\subsection{Model Statistics}
		Table~\ref{tab:stats} shows some statistics about the average training epoch times, numbers of trainable parameters, final model sizes and average forward times (i.e., the elapsed time for processing input data through a network to produce the next 5 time steps) for each DL model. Although TAU has the highest numbers of trainable parameters and a relatively larger model size, it still achieved the lowest training and forward times, due to its recurrence-free structure.
		
		\begin{table}[h]
			\centering
			\begin{tabular}{ccccccc}
				DL Model &
				\begin{tabular}[c]{@{}c@{}}Epoch Train\\ Time (min)\end{tabular} &
				\begin{tabular}[c]{@{}c@{}}Total Train\\ Time (h)\end{tabular} &
				\begin{tabular}[c]{@{}c@{}}Trainable\\ Parameters\end{tabular} &
				\begin{tabular}[c]{@{}c@{}}Model Size\\ (MB)\end{tabular} &
				\begin{tabular}[c]{@{}c@{}}Forward\\ Time (ms)\end{tabular}\\ \hline
				\textbf{ConvLSTM} & 6.44  & 10.73 & \textbf{4 M} & \textbf{13} & 60 \\
				\textbf{U-FNO}    & 10.98 & 18.30 & 6 M  & 25 & 79 & \\
				\textbf{TAU}      & \textbf{4.22} & \textbf{7.03} & 11 M & 135 & \textbf{33}
			\end{tabular}
			\caption{Performance statistics for all tested models.}
			\label{tab:stats}
		\end{table}
		
		Despite being the most lightweight model, ConvLSTM has a 50\% slower training time per epoch than TAU, also taking nearly twice as long for the forward operation. Finally, U-FNO achieved the slowest training and forward times, but the resulting model after the training process is about five times smaller than TAU and almost two times bigger compared to ConvLSTM.
		
		Analyzing the total runtime of each algorithm to produce a 100-step simulation, TAU takes an average time of 0.66 seconds, followed by ConvLSTM (1.2 seconds) and U-FNO (1.58 seconds). These represent a speedup with an order of magnitude between $10^3$ and $10^4$ when compared to the average time taken by GeoChemFoam (3 hours).
		
		\subsection{Iterative Prediction and Model Stacking Analysis}
		\label{sec:rec_pred}
		
\begin{figure}[h!]
			\centering
			\includegraphics[width=0.89\linewidth]{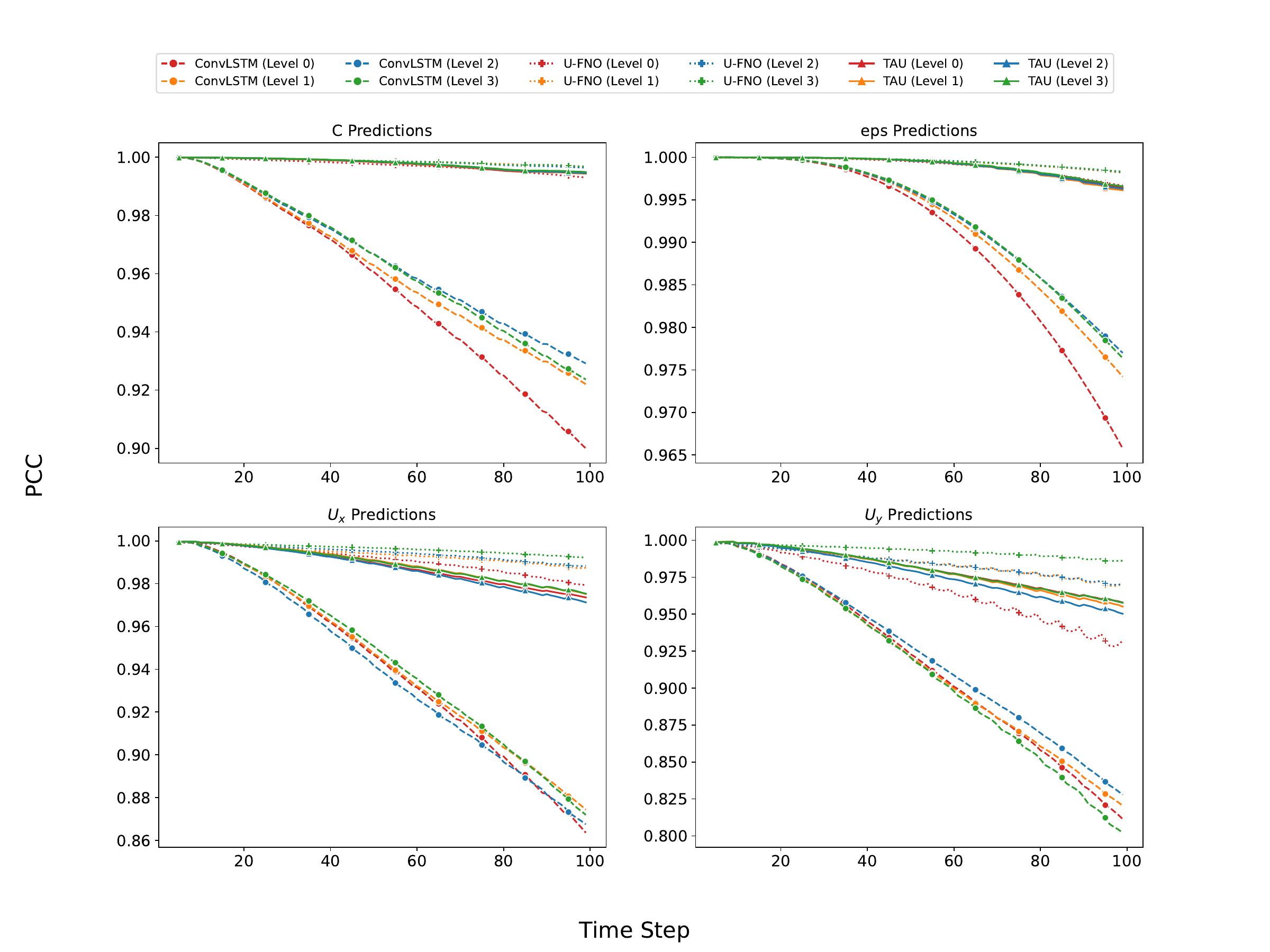}
			\caption{Average correlation scores of all samples from the training set for the iterative predictions produced by each algorithm.}
			\label{fig:results_train}
		\end{figure}

		Figure~\ref{fig:results_train} shows the average correlations of the iterative predictions on all training samples for all algorithms and correction levels. For all properties, it can be observed that both TAU and U-FNO are more robust to error accumulation, with the latter achieving slightly higher correlations at late time steps. With respect to each output property, the best results were achieved for $eps$ prediction, followed by the concentration $C$, and the flow directions $U_x$ and $U_y$.
		
		Considering the effects of model stacking, ConvLSTM achieved its best results at Level 2 correction (except for $U_x$ prediction). Still, it was not capable of performing better than the base networks (Level 0) of the other two algorithms. TAU showed no significant improvement when applying multi-level stacking for all cases, as evidenced by the fact that the correlations do not systematically increase over the correction levels. On the other hand, U-FNO showed a more consistent evolution from the base network to Level 3 in all scenarios, as the average correlations at the late time steps typically increase after each correction.
		
		\begin{figure}[h!]
			\centering
			\includegraphics[width=0.89\linewidth]{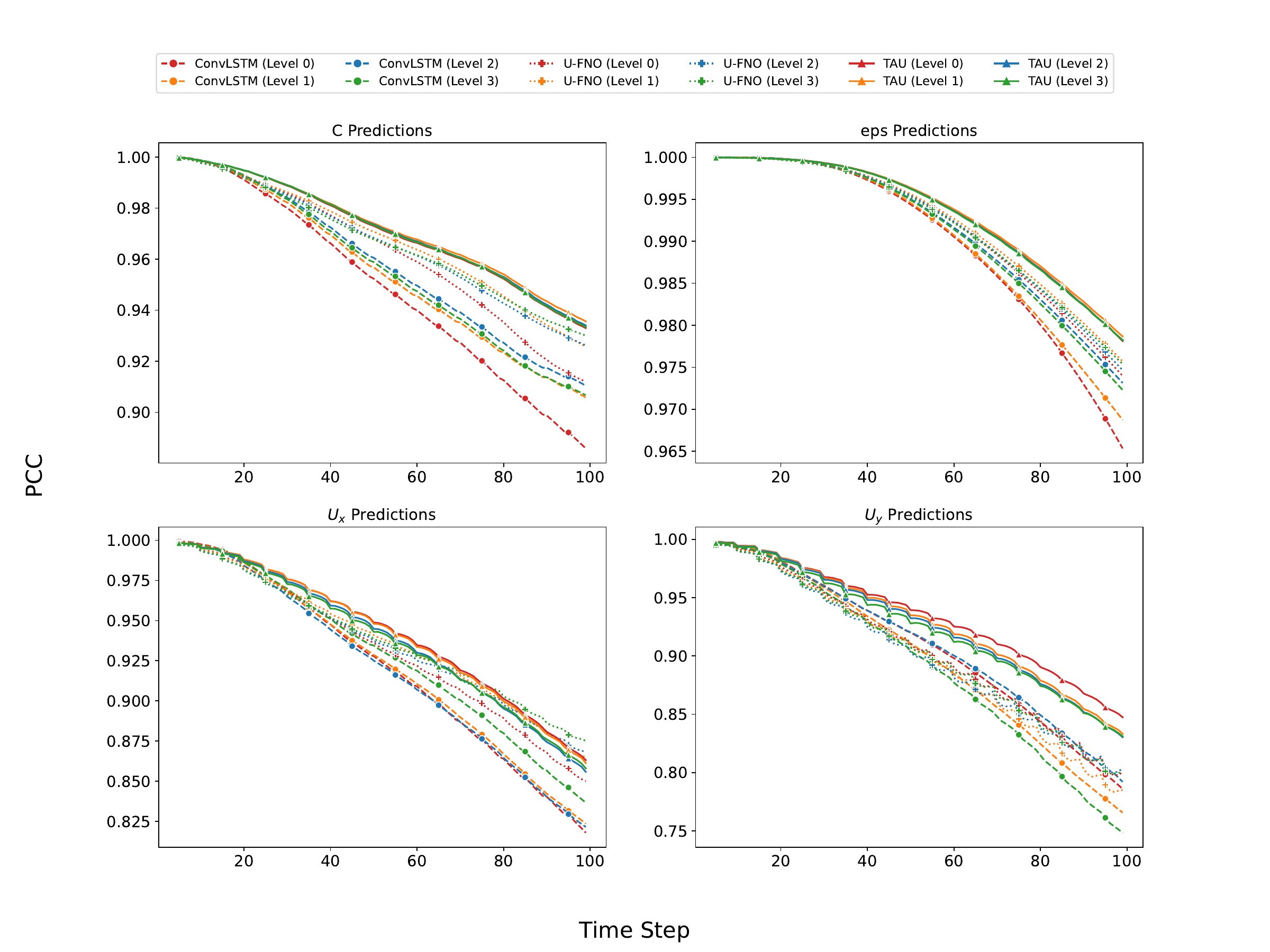}
			\caption{Average correlation scores of all samples from the validation set for the iterative predictions produced by each algorithm.}
			\label{fig:results_val}
		\end{figure}
		
		Figure~\ref{fig:results_val} illustrates the results for the validation set. The plots show that TAU achieved the best iterative predictions for all cases (except for $U_x$ prediction). Apart from ConvLSTM, the correlations had a reasonable drop when compared to those from the training data. However, this drop is much larger for the U-FNO, suggesting that this network is much more prone to over-fitting than the others.
		
		Regarding the performance evolution along the correction levels, ConvLSTM still yielded its best results on Level 2 network, except for $U_x$ prediction, in which Level 3 was more robust to error accumulation. However, it achieved the lowest correlations in almost all scenarios, being only superior to U-FNO in the prediction of $U_y$. The validation curves for U-FNO, unlike the results for the training set, did not show a consistent evolution over the levels. For instance, in the prediction of $C$, the Level 1 network achieved the best results until time step 80, where it was surpassed by Level 3. Moreover, Level 1 prevailed as the best correction network during all time steps of $eps$ prediction. Finally, the Level 1 network from TAU achieved the highest correlations for predictions of $C$ and $eps$, as well as for $U_x$ predictions until time step 50. Despite its better performance against ConvLSTM and U-FNO, the multi-level stacking strategy did not cause a significant improvement over the TAU Level 0 network.
		
		To ratify the robustness of our method by evaluating it with a different metric, we refer to~\ref{app:mse}.
		
		\subsection{Relevance of Engineered Features}
		
\begin{figure}[h]
			\centering
			\includegraphics[width=\linewidth]{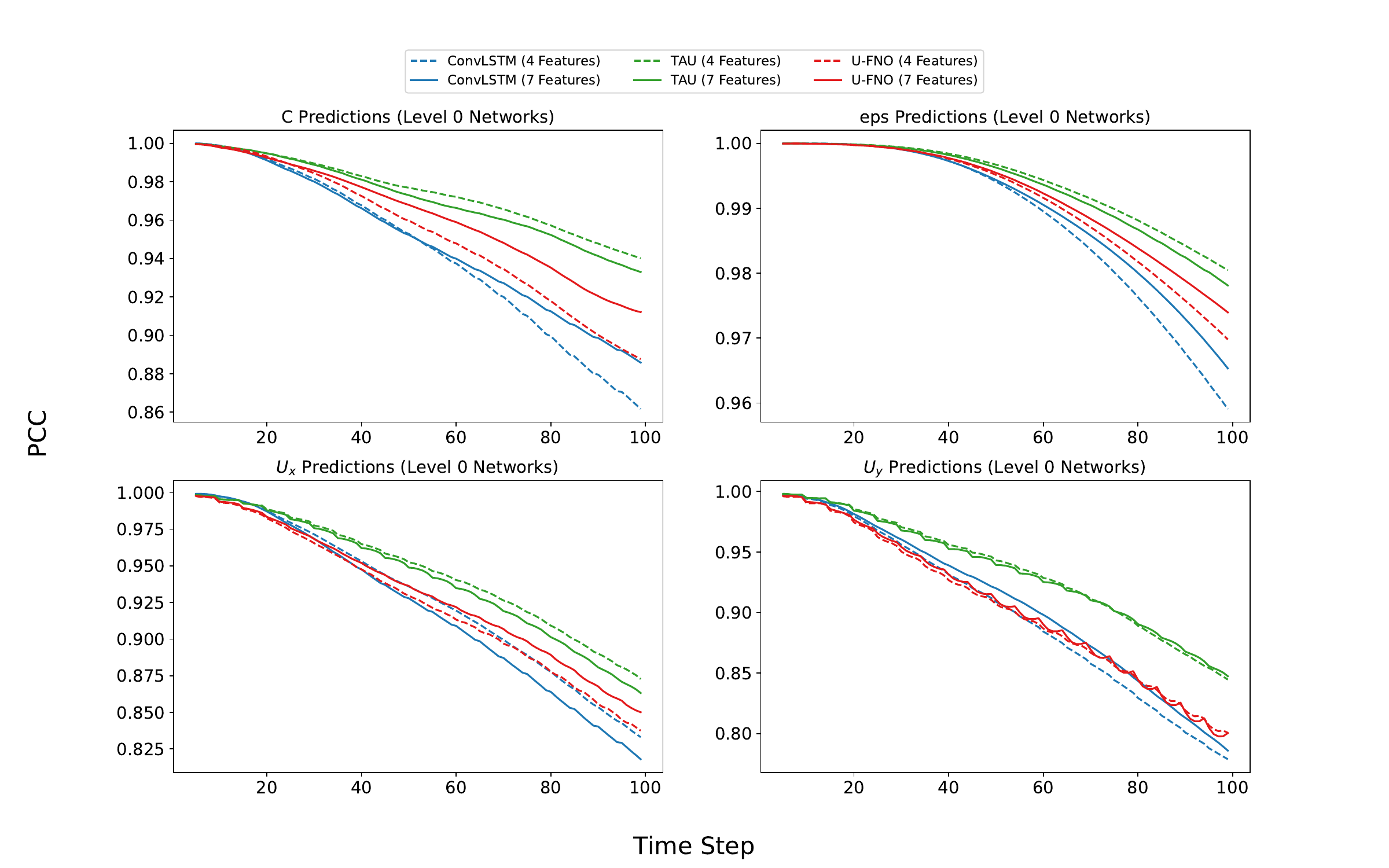}
			\caption{Average correlations for all models in the validation set and each output property, comparing the performances of all Level 0 networks with and without the engineered features, respectively represented by solid and dashed lines.}
			\label{fig:eng_feat}
		\end{figure}

		Figure~\ref{fig:eng_feat} illustrates the effects of including the three engineered features described in Section~\ref{sec:dataset} along with the four original input properties. To provide a clear visualization of these effects, we will only discuss the results with respect to Level 0 networks. From the plots, we can notice that for both ConvLSTM and U-FNO, there is a performance drop when transitioning from 7 to 4 input features, except for $U_x$ predictions for ConvLSTM. Moreover, U-FNO with 4 input features achieved better results than ConvLSTM with 7 input features.
		
		Conversely, TAU yielded slightly better performances without the engineered features, also achieving superior results than the other two methods. However, a further analysis with respect to its internal parameters (including the $\alpha$ term of the loss function) must be conducted to confirm whether the engineered features are actually relevant. Nevertheless, the plots demonstrate TAU's capability to understand the evolution of the reactive dissolution process with a smaller set of input features. 
		
		\subsection{Qualitative Analysis}
		\label{subsec:qual}
		
\begin{figure}[hbt!]
			\centering
			\begin{minipage}[b]{0.495\textwidth}
				\centering
				\includegraphics[width=\textwidth]{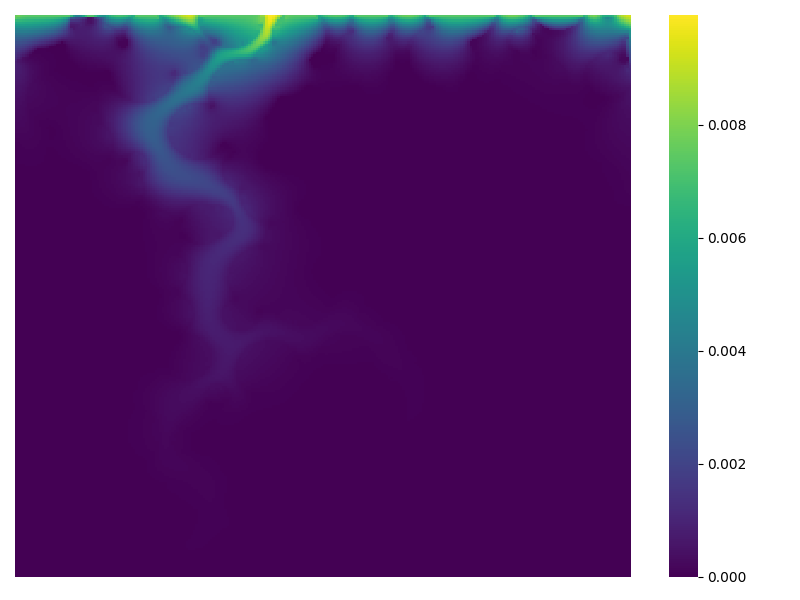}
				\subcaption{True $C$ (Time Step 0)}
				\label{fig:true_c_step0}
			\end{minipage}
			\hfill
			\begin{minipage}[b]{0.495\textwidth}
				\centering
				\includegraphics[width=\textwidth]{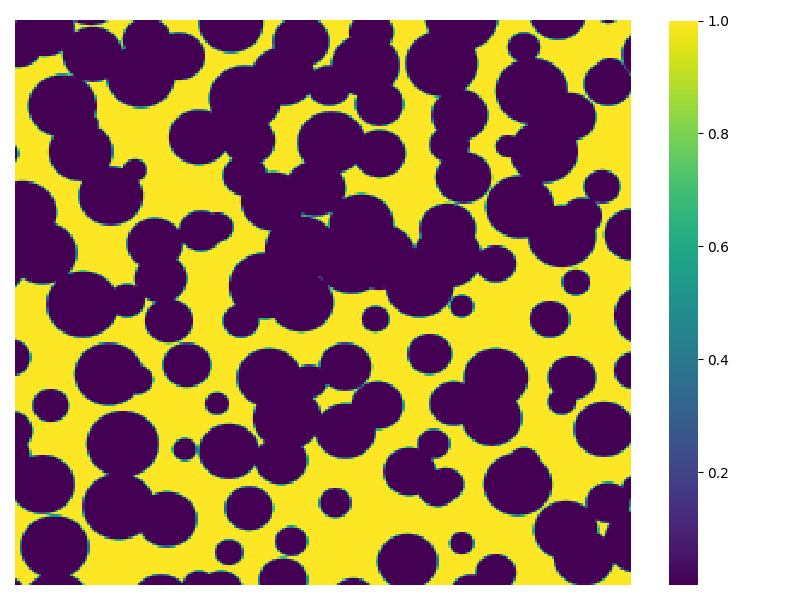}
				\subcaption{True $eps$ (Time Step 0)}
				\label{fig:true_eps_step0}
			\end{minipage} \\
			
			\begin{minipage}[b]{0.495\textwidth}
				\centering
				\includegraphics[width=\textwidth]{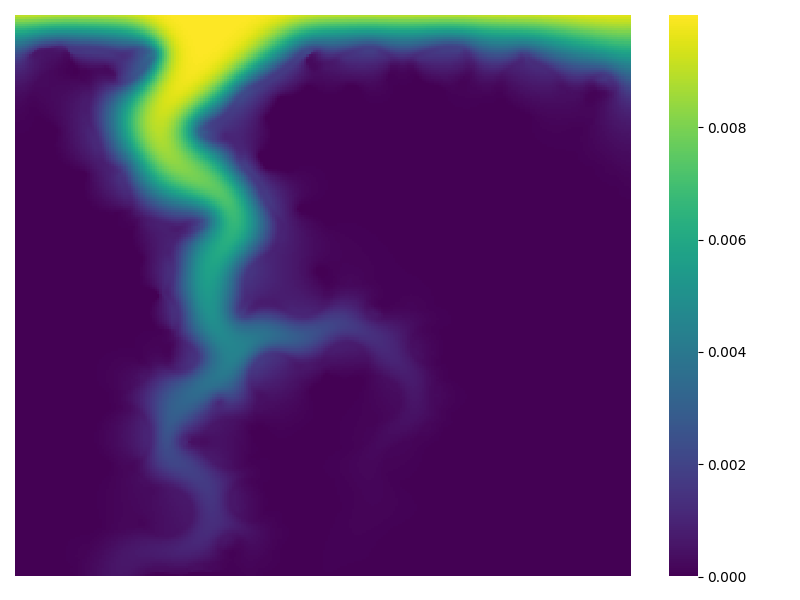}
				\subcaption{True $C$ (Time Step 100)}
				\label{fig:true_c_step100}
			\end{minipage}
			\hfill
			\begin{minipage}[b]{0.495\textwidth}
				\centering
				\includegraphics[width=\textwidth]{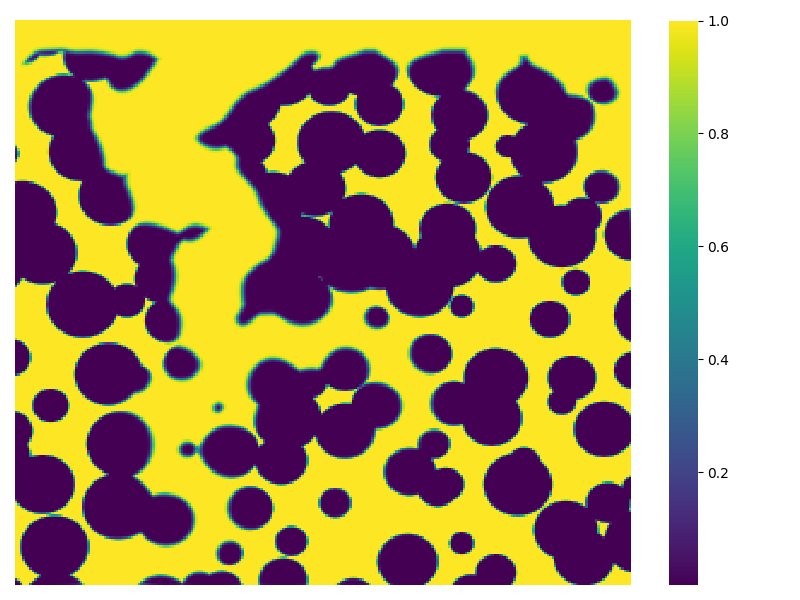}
				\subcaption{True $eps$ (Time Step 100)}
				\label{fig:true_eps_step100}
			\end{minipage}
			\caption{Ground truth maps at time steps 0 and 100 for $C$ and $eps$.}
			\label{fig:gt}    
		\end{figure}

		In this section, we will conduct a qualitative analysis on a sample from the validation set, comprising a particular pore geometry. Herein, we will consider the predictions at the last time step (100) after several iterative steps are performed for each algorithm. To simplify our analysis, we only show the predictions for $C$ and $eps$ fields. Figure~\ref{fig:gt} shows the ground truth maps for those two properties at time steps 0 and 100.
		
\begin{figure}[hbt!]
			\begin{minipage}[b]{0.325\textwidth}
				\centering
				\includegraphics[width=\textwidth]{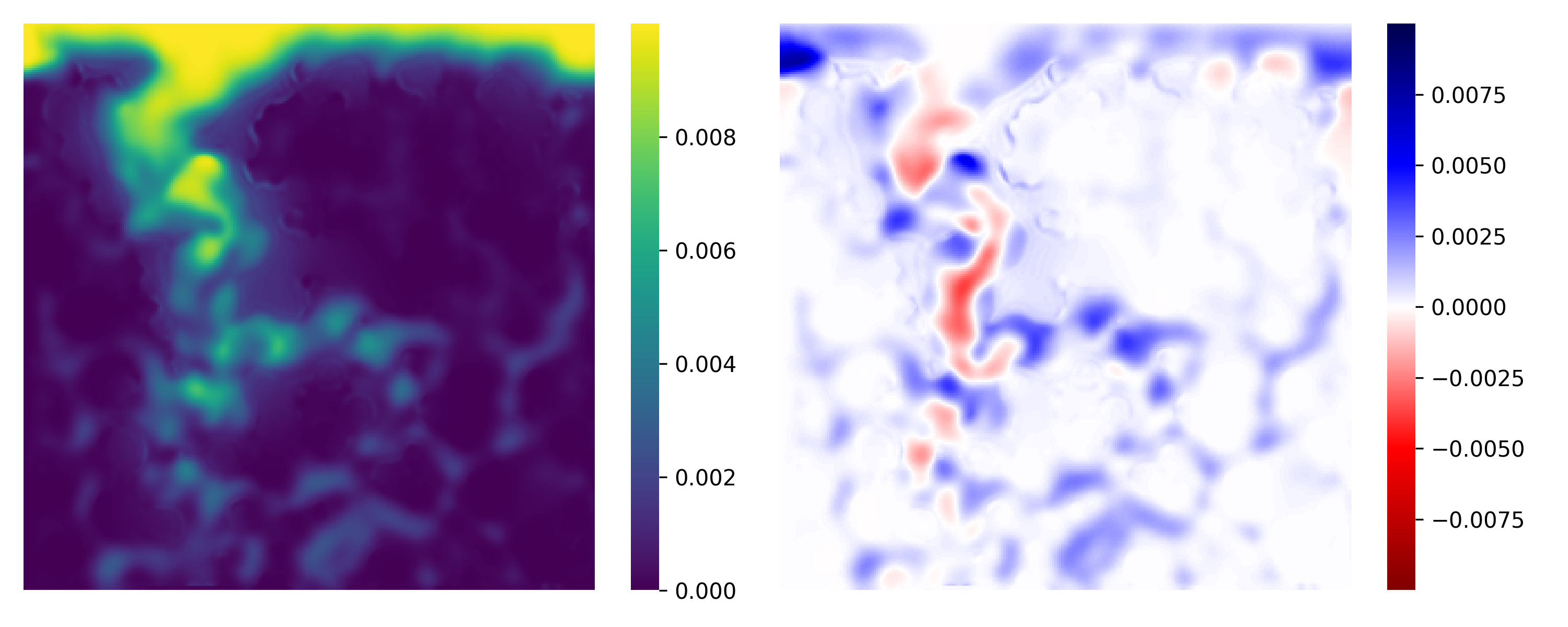}
				\subcaption{ConvLSTM Level 0\\PCC = 0.944}
				\label{fig:convlstm_lv0_c_1}
			\end{minipage}
			\hfill
			\begin{minipage}[b]{0.325\textwidth}
				\centering
				\includegraphics[width=\textwidth]{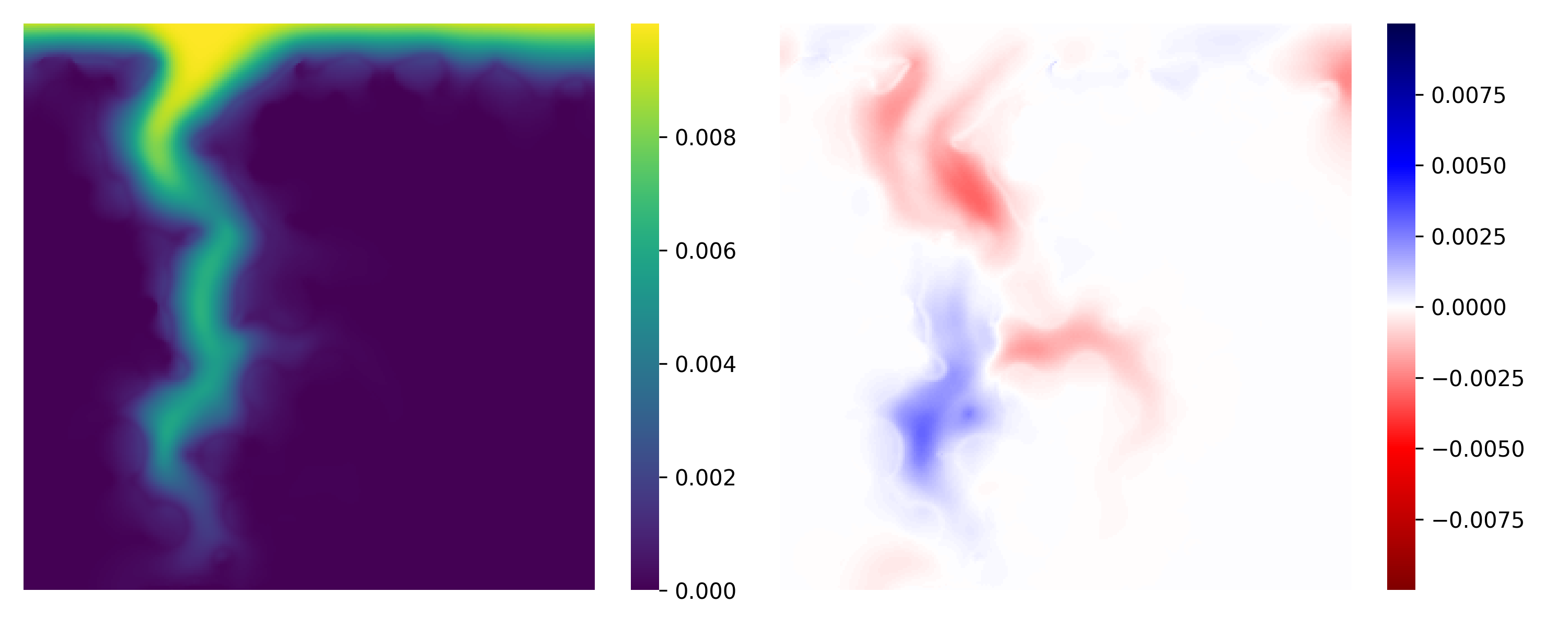}
				\subcaption{U-FNO Level 0\\PCC = 0.975}
				\label{fig:ufno_lv0_c_1}
			\end{minipage}
			\hfill
			\begin{minipage}[b]{0.325\textwidth}
				\centering
				\includegraphics[width=\textwidth]{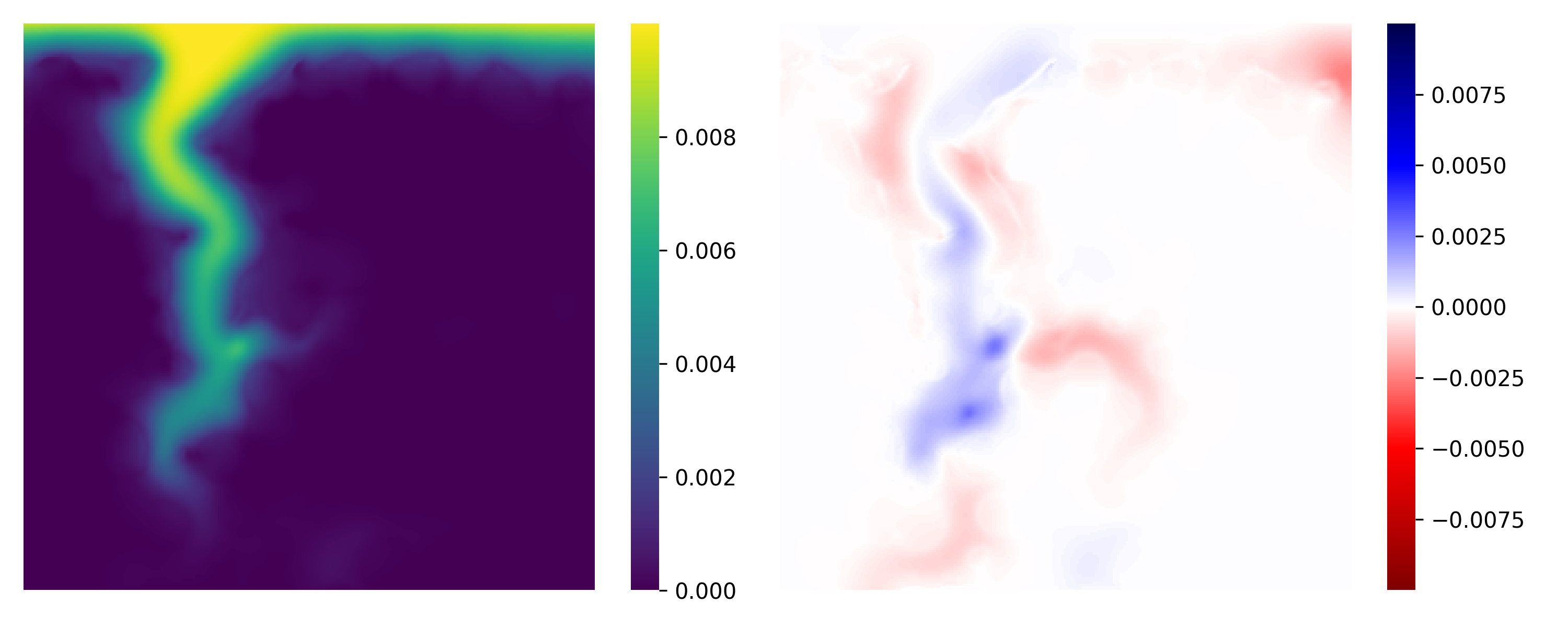}
				\subcaption{TAU Level 0\\\textbf{PCC = 0.987}}
				\label{fig:tau_lv0_c_1}
			\end{minipage} \\
			\begin{minipage}[b]{0.325\textwidth}
				\centering
				\includegraphics[width=\textwidth]{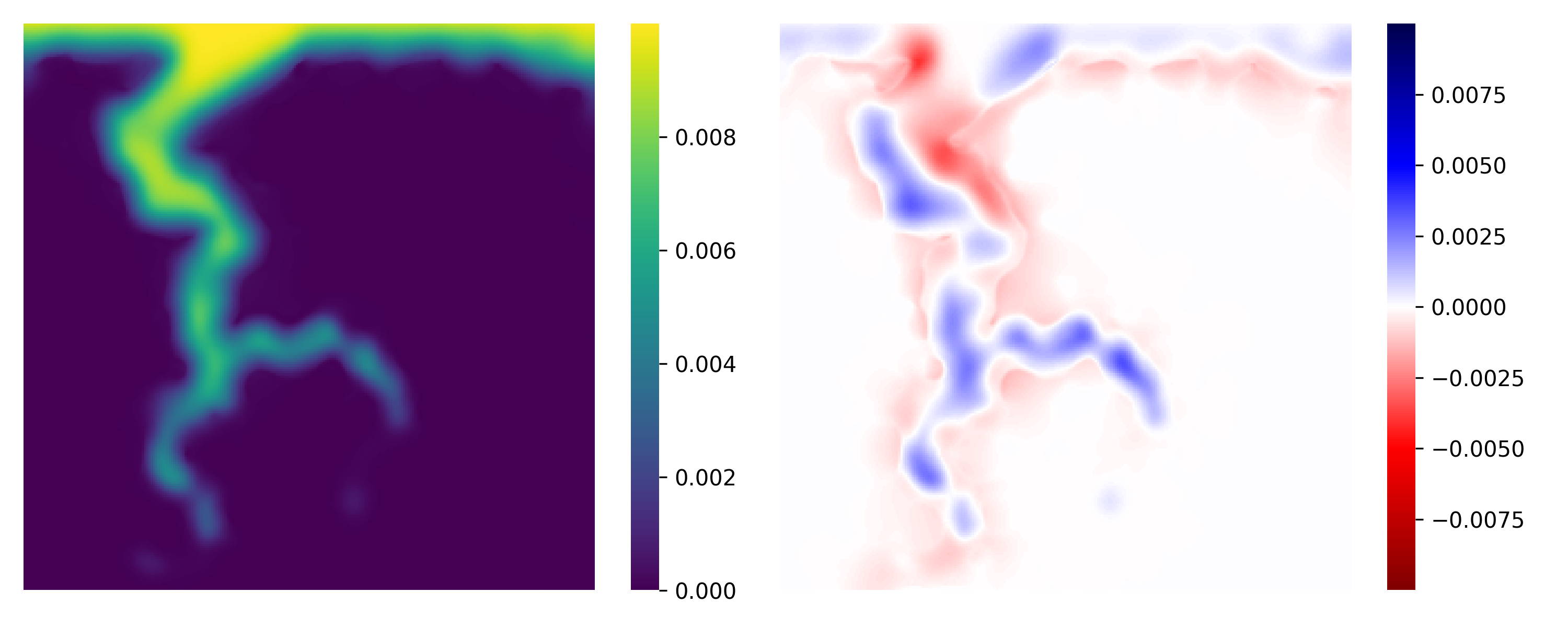}
				\subcaption{ConvLSTM Level 1\\PCC = 0.969}
				\label{fig:convlstm_lv1_c_1}
			\end{minipage}
			\hfill
			\begin{minipage}[b]{0.325\textwidth}
				\centering
				\includegraphics[width=\textwidth]{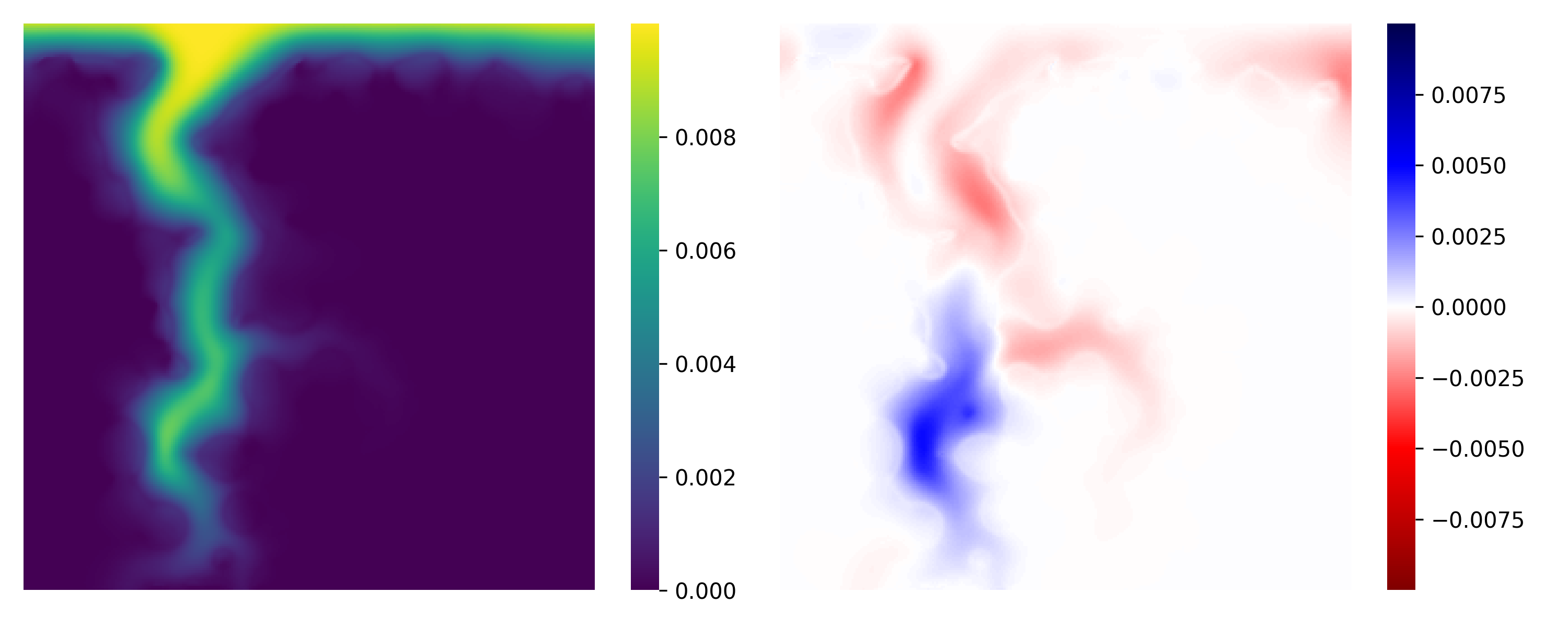}
				\subcaption{U-FNO Level 1\\PCC = 0.956}
				\label{fig:ufno_lv1_c_1}
			\end{minipage}
			\hfill
			\begin{minipage}[b]{0.325\textwidth}
				\centering
				\includegraphics[width=\textwidth]{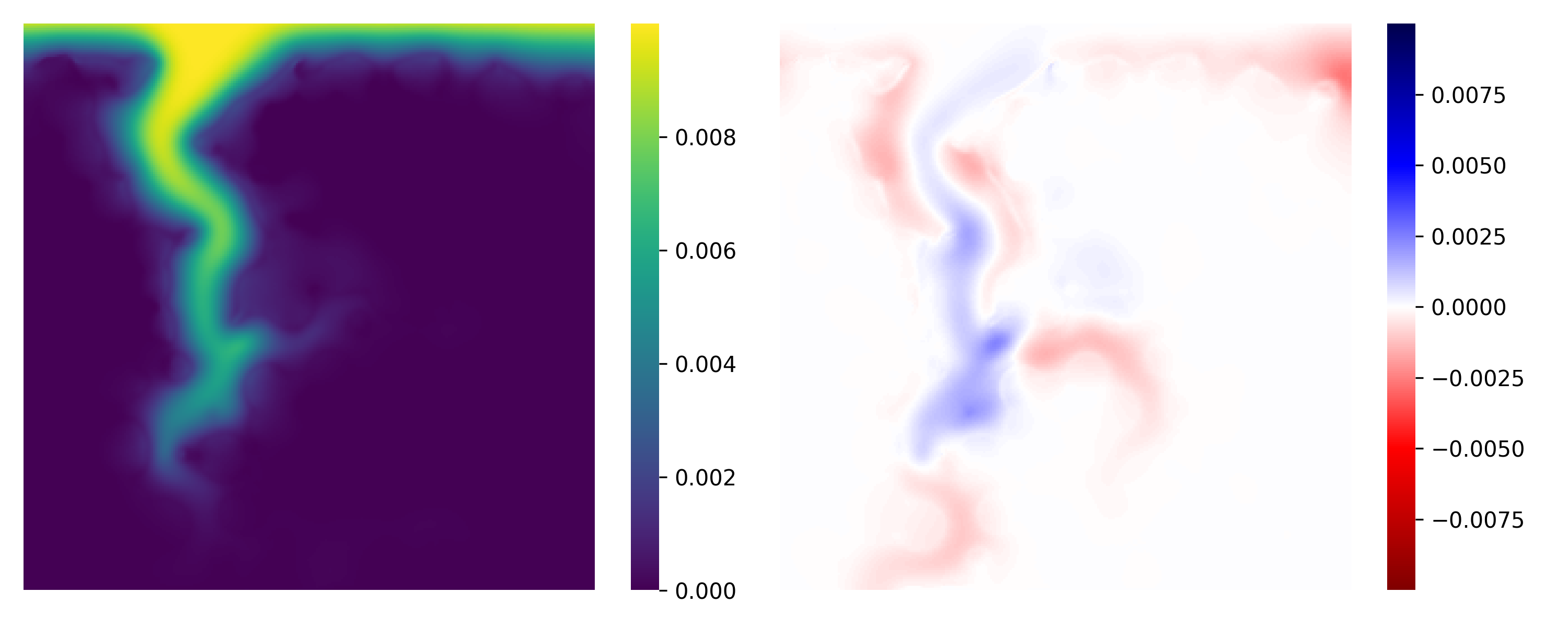}
				\subcaption{TAU Level 1\\\textbf{PCC = 0.987}}
				\label{fig:tau_lv1_c_1}
			\end{minipage} \\
			\begin{minipage}[b]{0.325\textwidth}
				\centering
				\includegraphics[width=\textwidth]{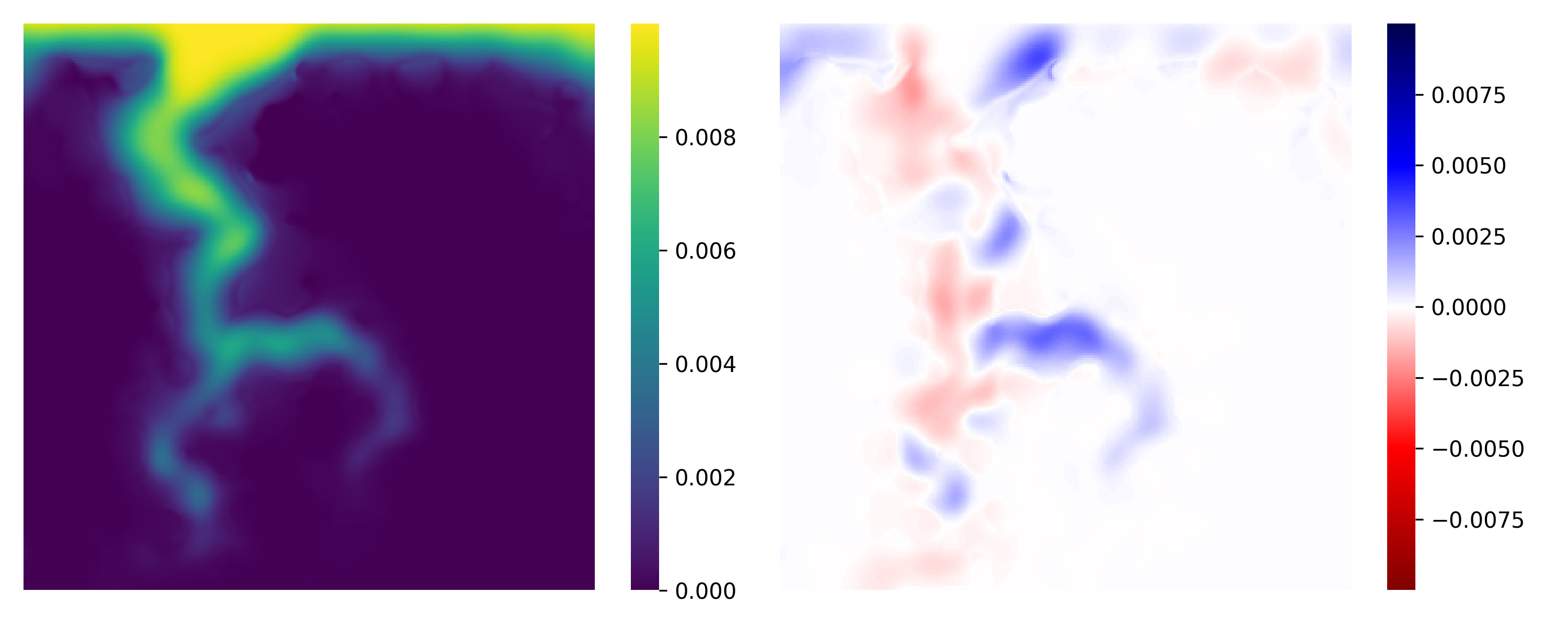}
				\subcaption{ConvLSTM Level 2\\PCC = 0.979}
				\label{fig:convlstm_lv2_c_1}
			\end{minipage}
			\hfill
			\begin{minipage}[b]{0.325\textwidth}
				\centering
				\includegraphics[width=\textwidth]{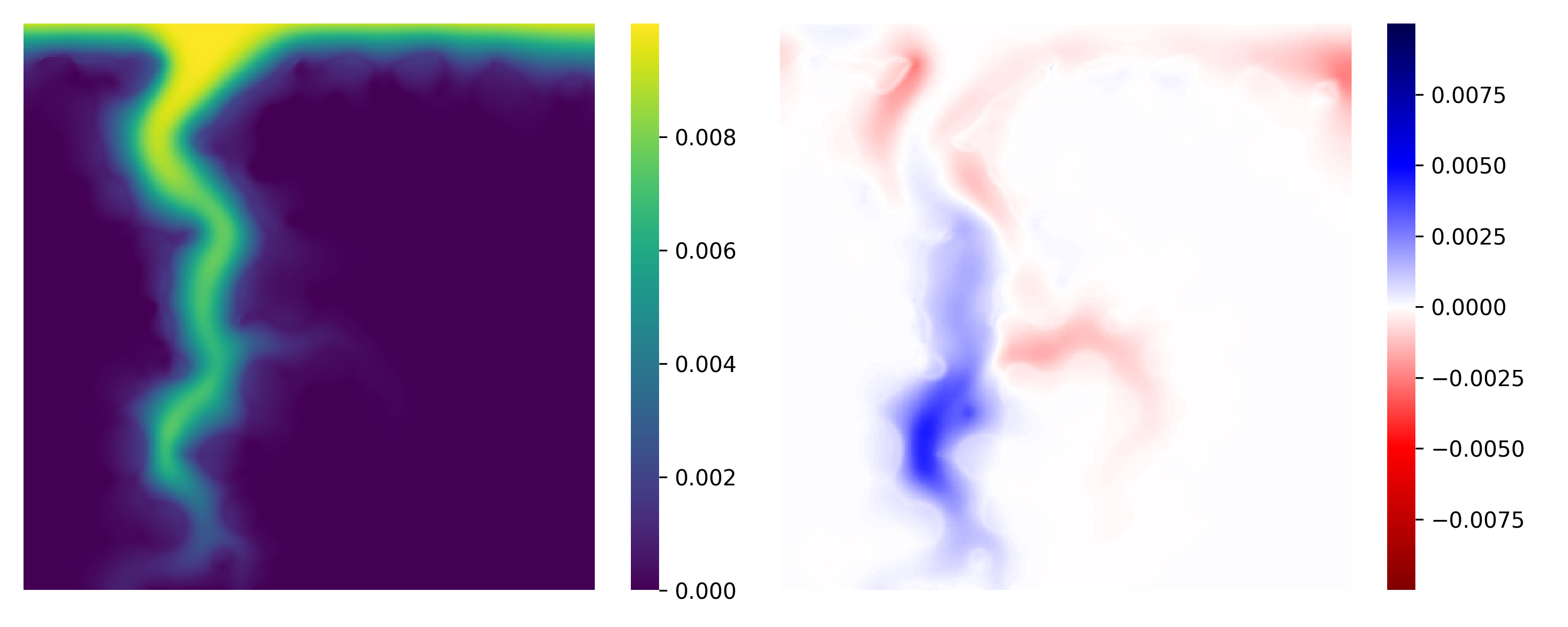}
				\subcaption{U-FNO Level 2\\PCC = 0.965}
				\label{fig:ufno_lv2_c_1}
			\end{minipage}
			\hfill
			\begin{minipage}[b]{0.325\textwidth}
				\centering
				\includegraphics[width=\textwidth]{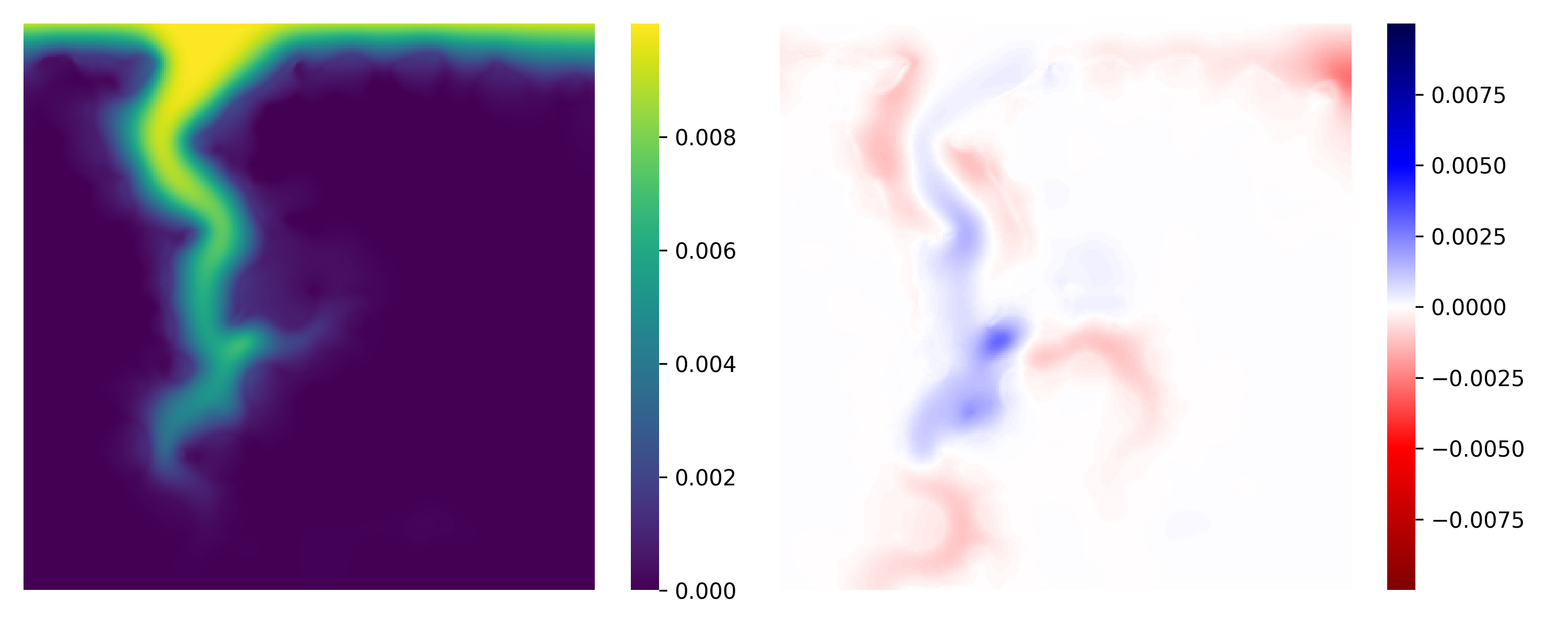}
				\subcaption{TAU Level 2\\\textbf{PCC = 0.987}}
				\label{fig:tau_lv2_c_1}
			\end{minipage} \\
			\begin{minipage}[b]{0.325\textwidth}
				\centering
				\includegraphics[width=\textwidth]{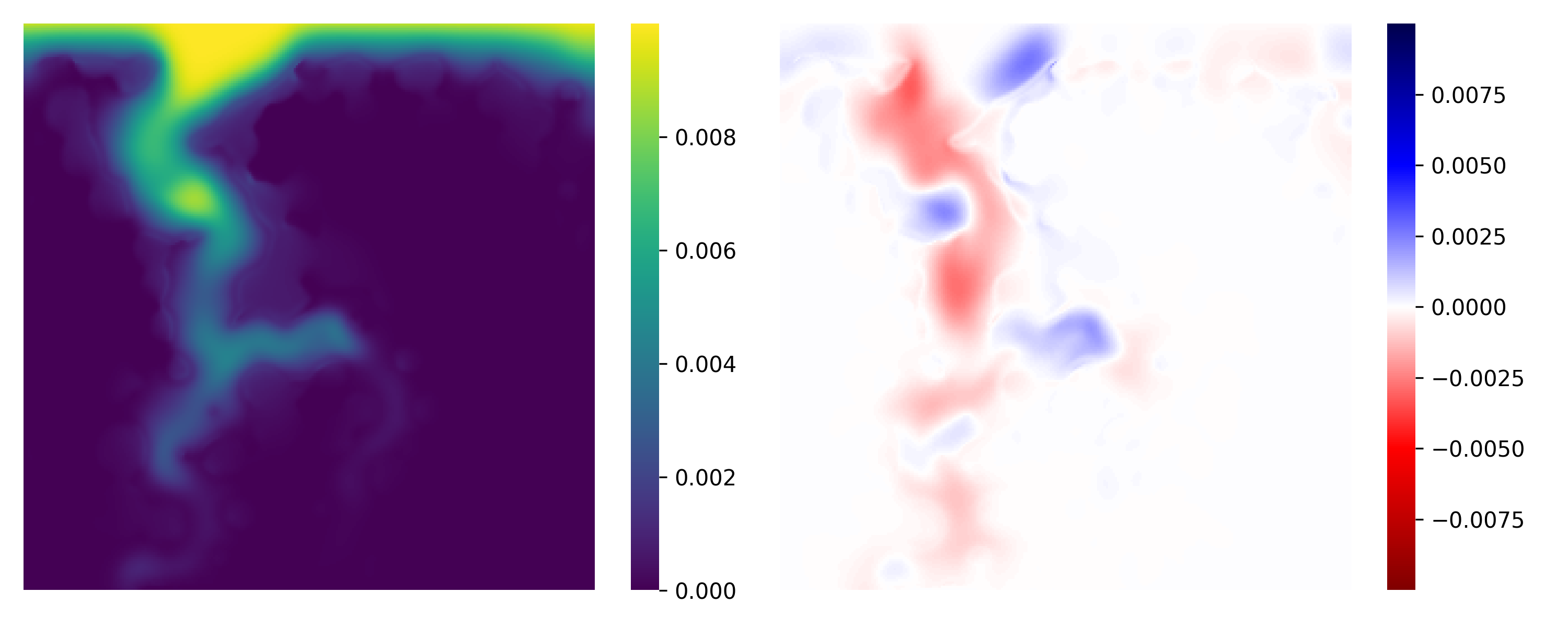}
				\subcaption{ConvLSTM Level 3\\PCC = 0.976}
				\label{fig:convlstm_lv3_c_1}
			\end{minipage}
			\hfill
			\begin{minipage}[b]{0.325\textwidth}
				\centering
				\includegraphics[width=\textwidth]{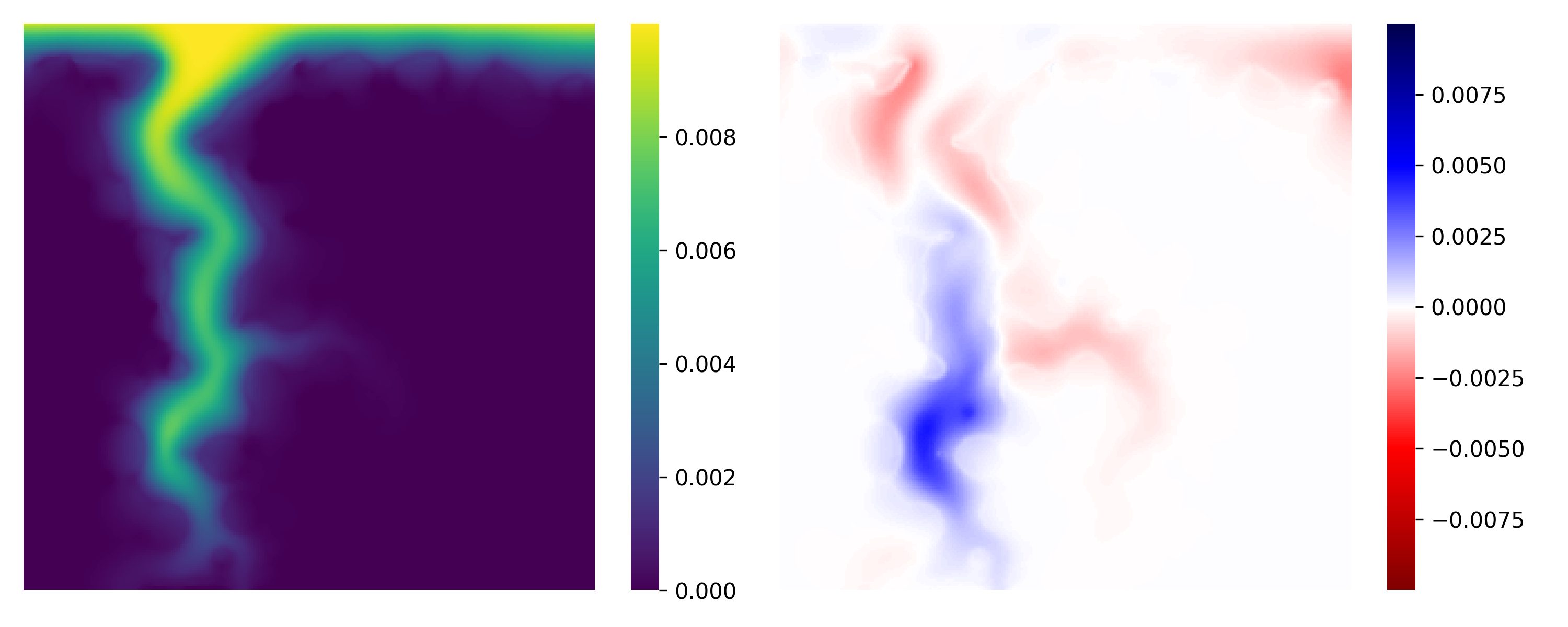}
				\subcaption{U-FNO Level 3\\PCC = 0.961}
				\label{fig:ufno_lv3_c_1}
			\end{minipage}
			\hfill
			\begin{minipage}[b]{0.325\textwidth}
				\centering
				\includegraphics[width=\textwidth]{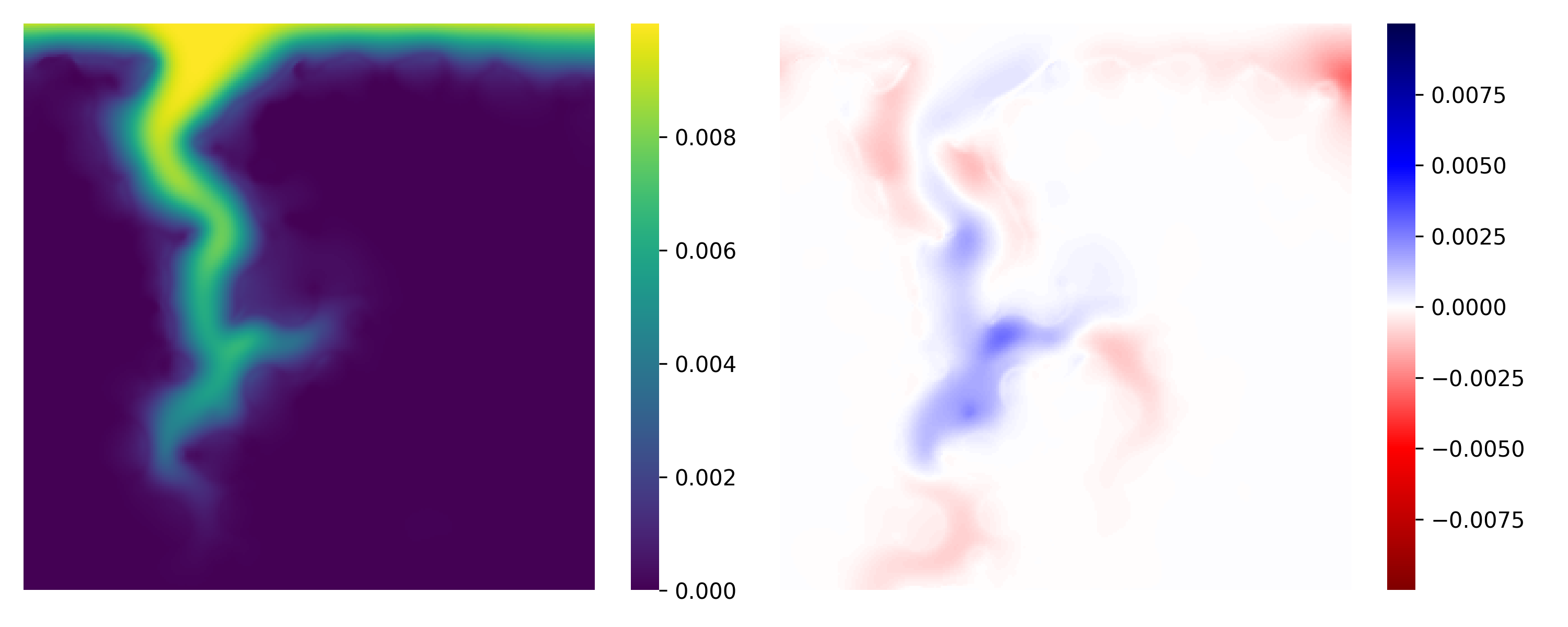}
				\subcaption{TAU Level 3\\\textbf{PCC = 0.986}}
				\label{fig:tau_lv3_c_1}
			\end{minipage}
			\caption{Predictions of $C$ at time step 100 for a case from the validation set, along with their respective PCC scores and difference maps to ground truth (Figure~\ref{fig:true_c_step100}). Best results for each network level are in bold.}
			\label{fig:c_preds_case1}
		\end{figure}

		The predictions of $C$ for all algorithms and all network levels are displayed in Figure~\ref{fig:c_preds_case1}, along with their respective difference maps to the ground truth ($\hat{Y} - Y$). We can notice a clear evolution on the ConvLSTM results, as the results are Level 0 contain too much noise, which is mostly mitigated over the subsequent levels. However, there is a slight decrease in the PCC score from Level 2 to Level 3. Regarding U-FNO, the best results were achieved at Level 0, which yielded a PCC of 0.004 lower than ConvLSTM Level 2. Moreover, all levels produced similar shapes of dissolution channel, only missing the right branching at the bottom-center part of the map. Finally, although not having benefitted from the multi-level stacking approach, TAU achieved the highest scores for all levels, where even its Level 0 network performed better than any other level from both ConvLSTM and U-FNO. Looking at the difference maps, we can also notice a smaller range of errors compared to the other two algorithms. With respect to the dissolution shape, it was able to capture some of the right branching, but not as much as ConvLSTM did.
		
\begin{figure}[hbt!]
			\begin{minipage}[b]{0.325\textwidth}
				\centering
				\includegraphics[width=\textwidth]{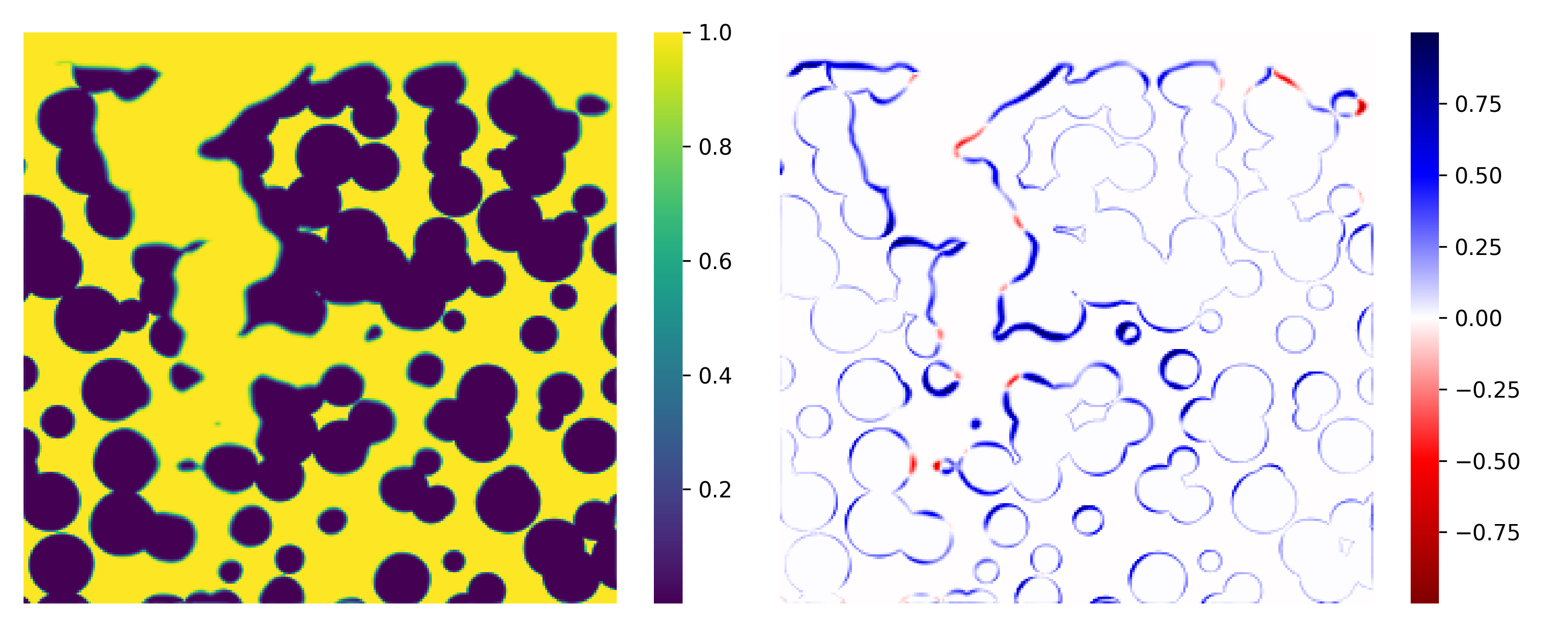}
				\subcaption{ConvLSTM Level 0\\PCC = 0.979}
				\label{fig:convlstm_lv0_eps_1}
			\end{minipage}
			\hfill
			\begin{minipage}[b]{0.325\textwidth}
				\centering
				\includegraphics[width=\textwidth]{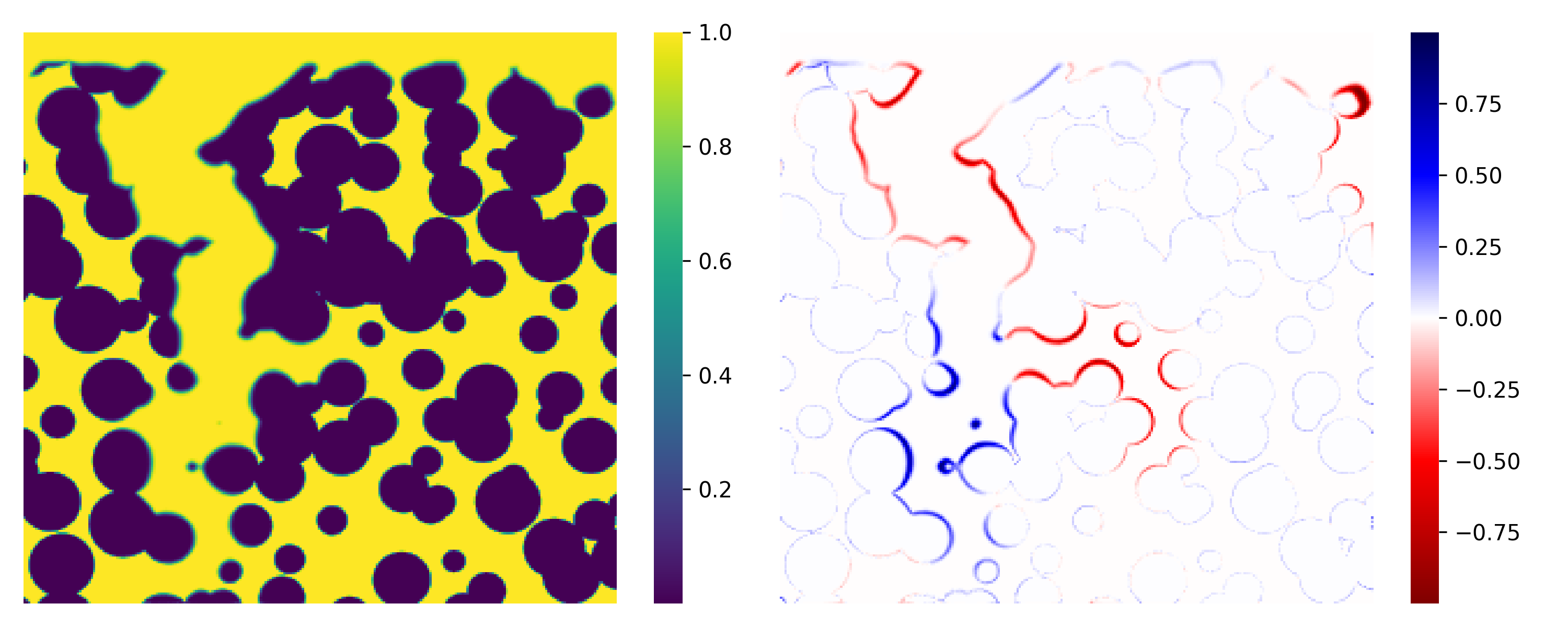}
				\subcaption{U-FNO Level 0\\PCC = 0.989}
				\label{fig:ufno_lv0_eps_1}
			\end{minipage}
			\hfill
			\begin{minipage}[b]{0.325\textwidth}
				\centering
				\includegraphics[width=\textwidth]{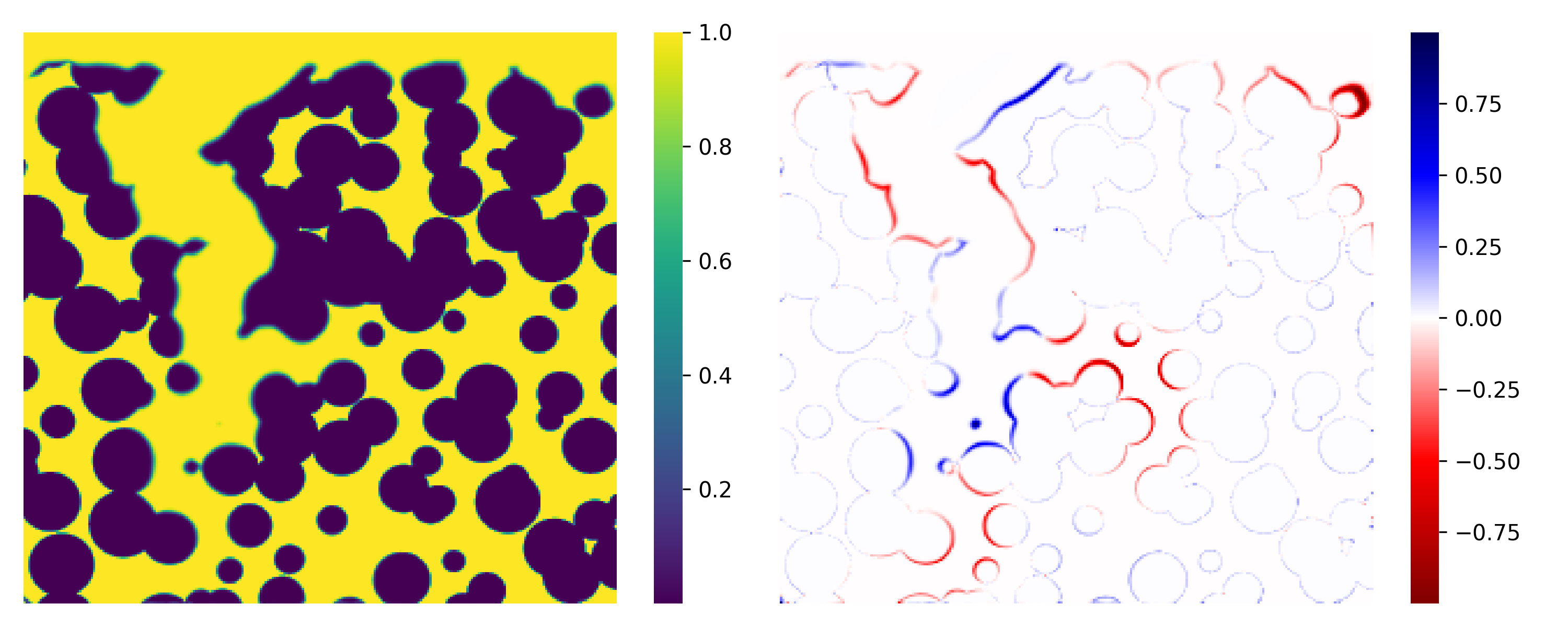}
				\subcaption{TAU Level 0\\\textbf{PCC = 0.992}}
				\label{fig:tau_lv0_eps_1}
			\end{minipage} \\
			\begin{minipage}[b]{0.325\textwidth}
				\centering
				\includegraphics[width=\textwidth]{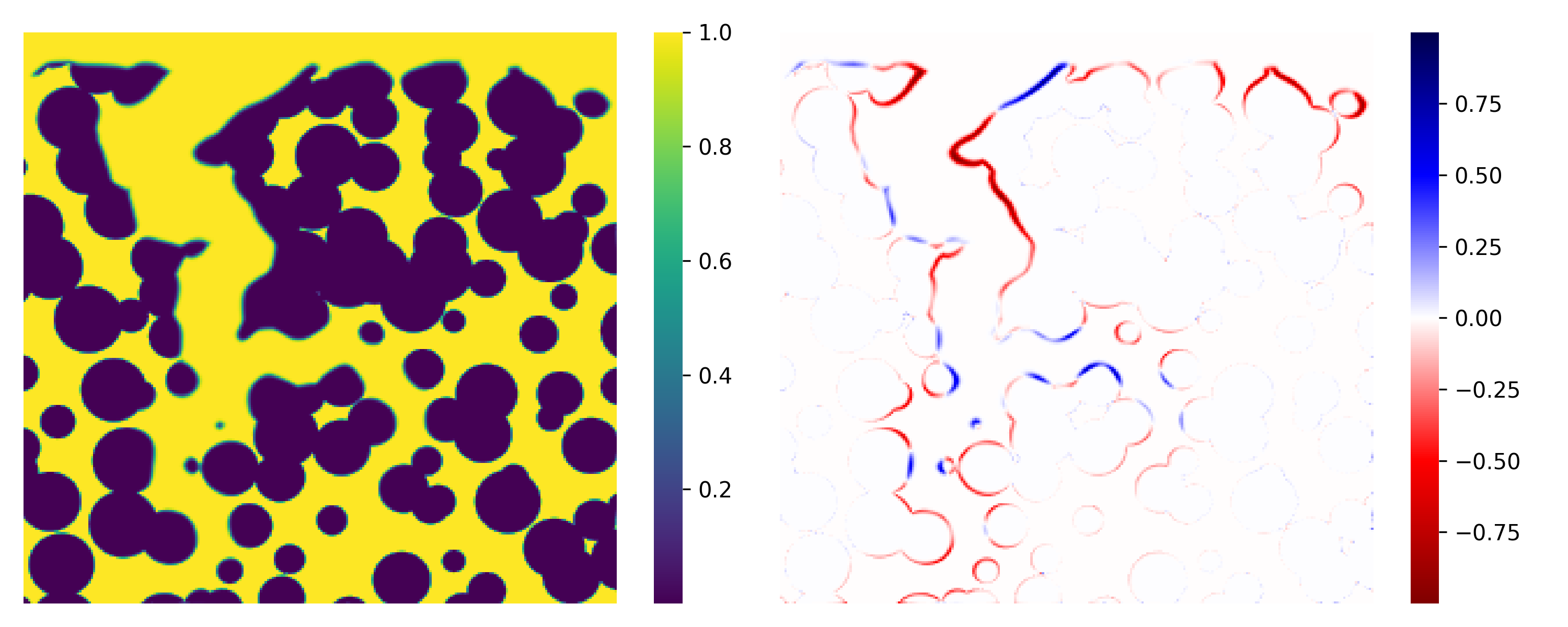}
				\subcaption{ConvLSTM Level 1\\PCC = 0.990}
				\label{fig:convlstm_lv1_eps_1}
			\end{minipage}
			\hfill
			\begin{minipage}[b]{0.325\textwidth}
				\centering
				\includegraphics[width=\textwidth]{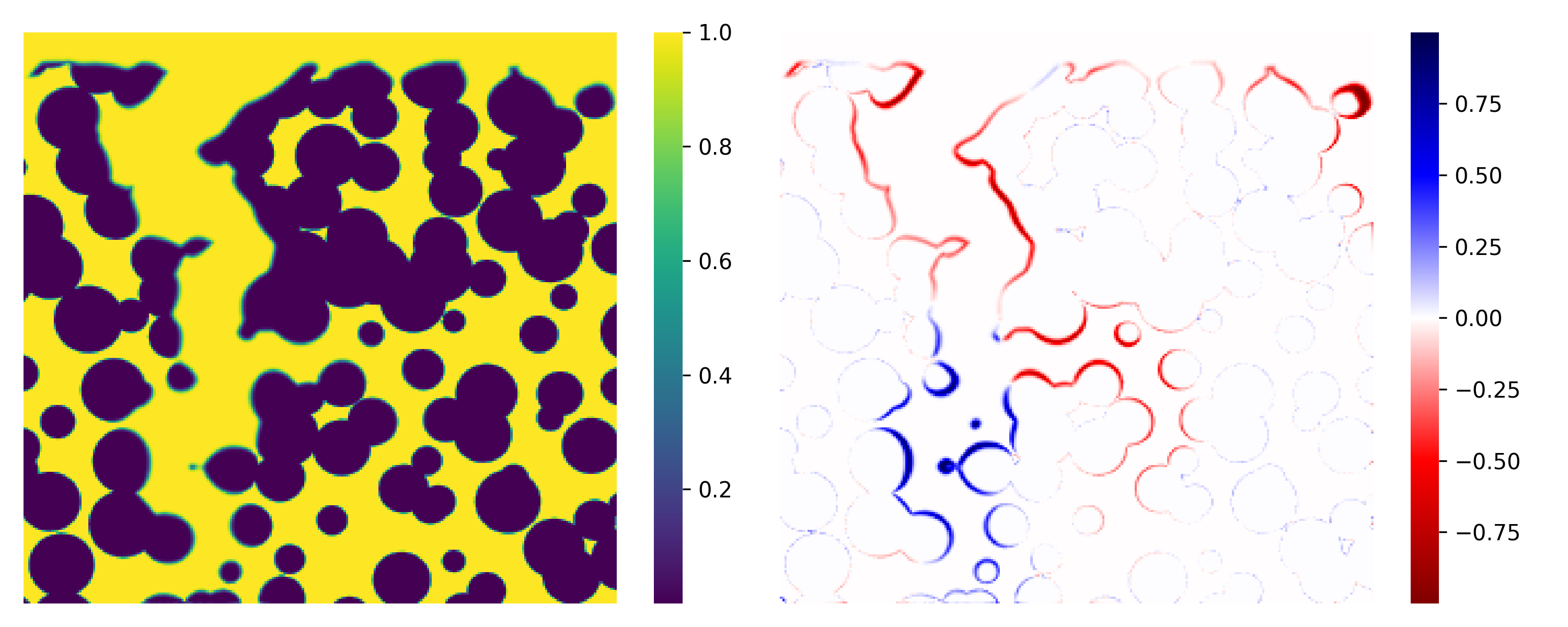}
				\subcaption{U-FNO Level 1\\PCC = 0.985}
				\label{fig:ufno_lv1_eps_1}
			\end{minipage}
			\hfill
			\begin{minipage}[b]{0.325\textwidth}
				\centering
				\includegraphics[width=\textwidth]{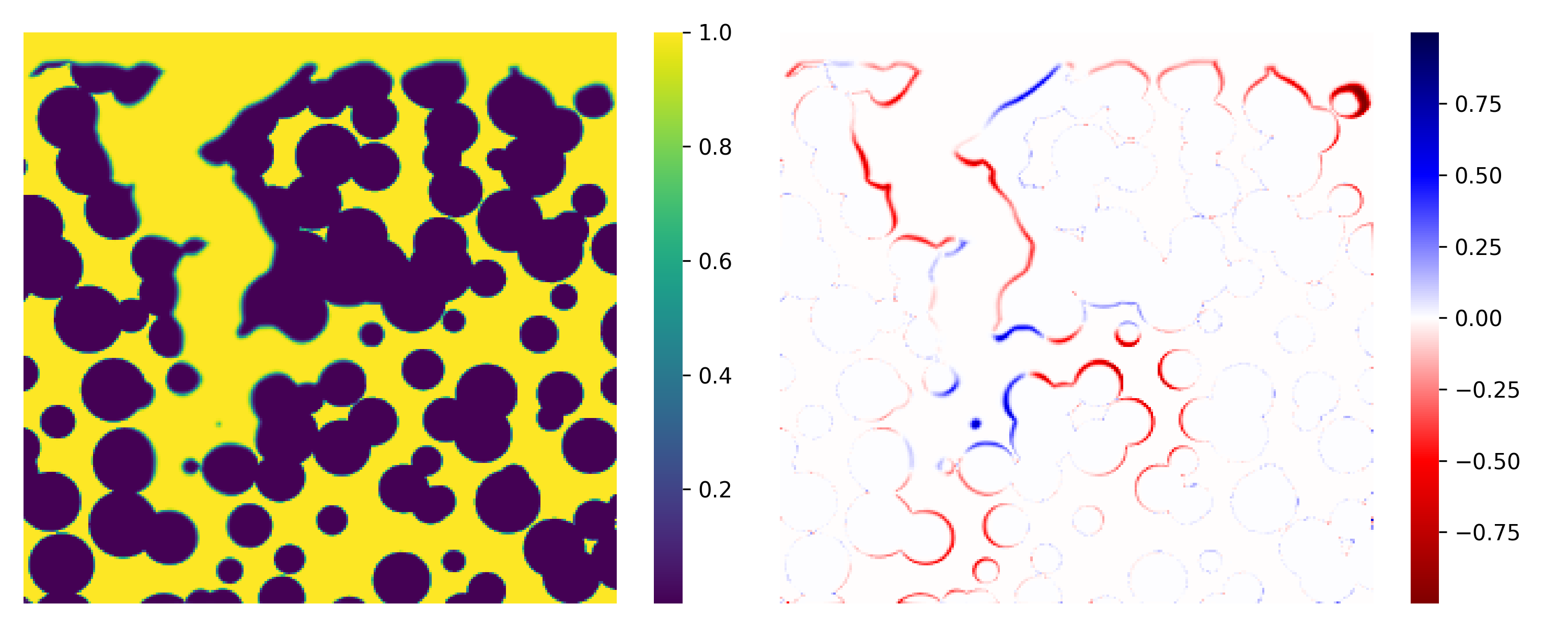}
				\subcaption{TAU Level 1\\\textbf{PCC = 0.991}}
				\label{fig:tau_lv1_eps_1}
			\end{minipage} \\
			\begin{minipage}[b]{0.325\textwidth}
				\centering
				\includegraphics[width=\textwidth]{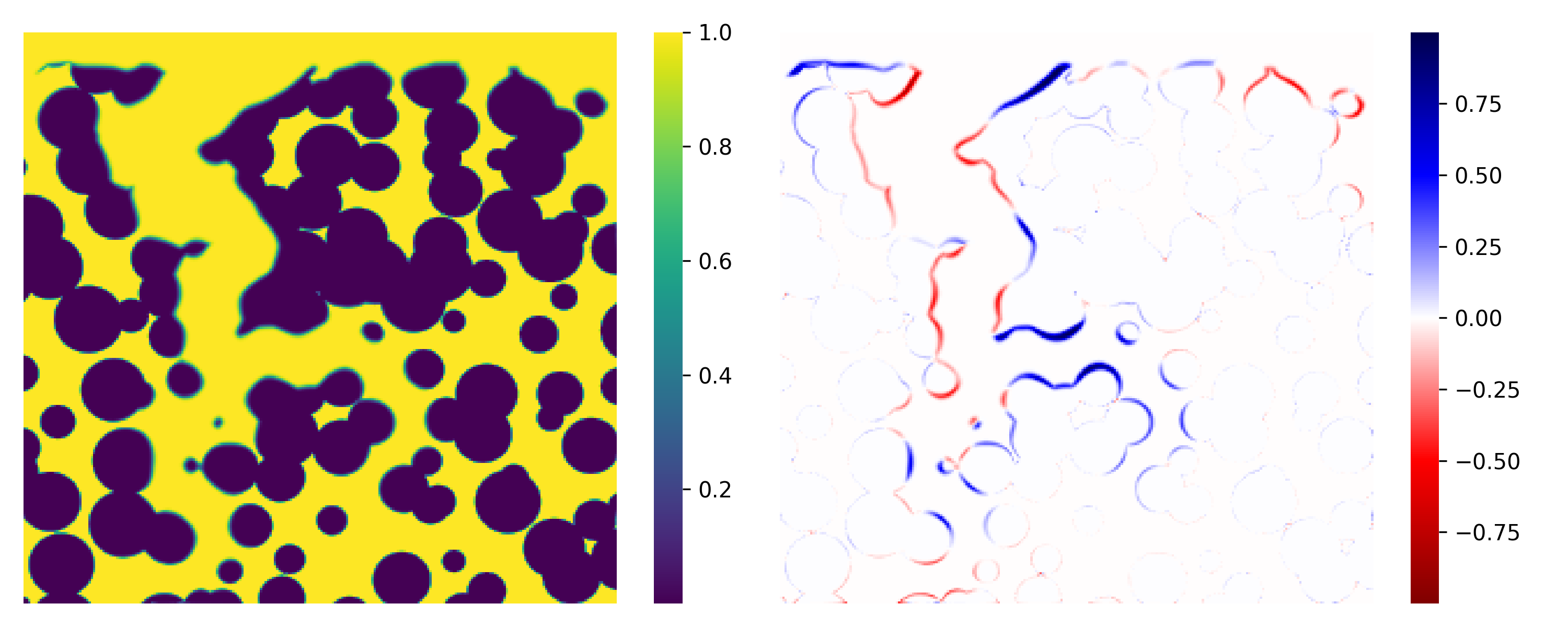}
				\subcaption{ConvLSTM Level 2\\PCC = 0.991}
				\label{fig:convlstm_lv2_eps_1}
			\end{minipage}
			\hfill
			\begin{minipage}[b]{0.325\textwidth}
				\centering
				\includegraphics[width=\textwidth]{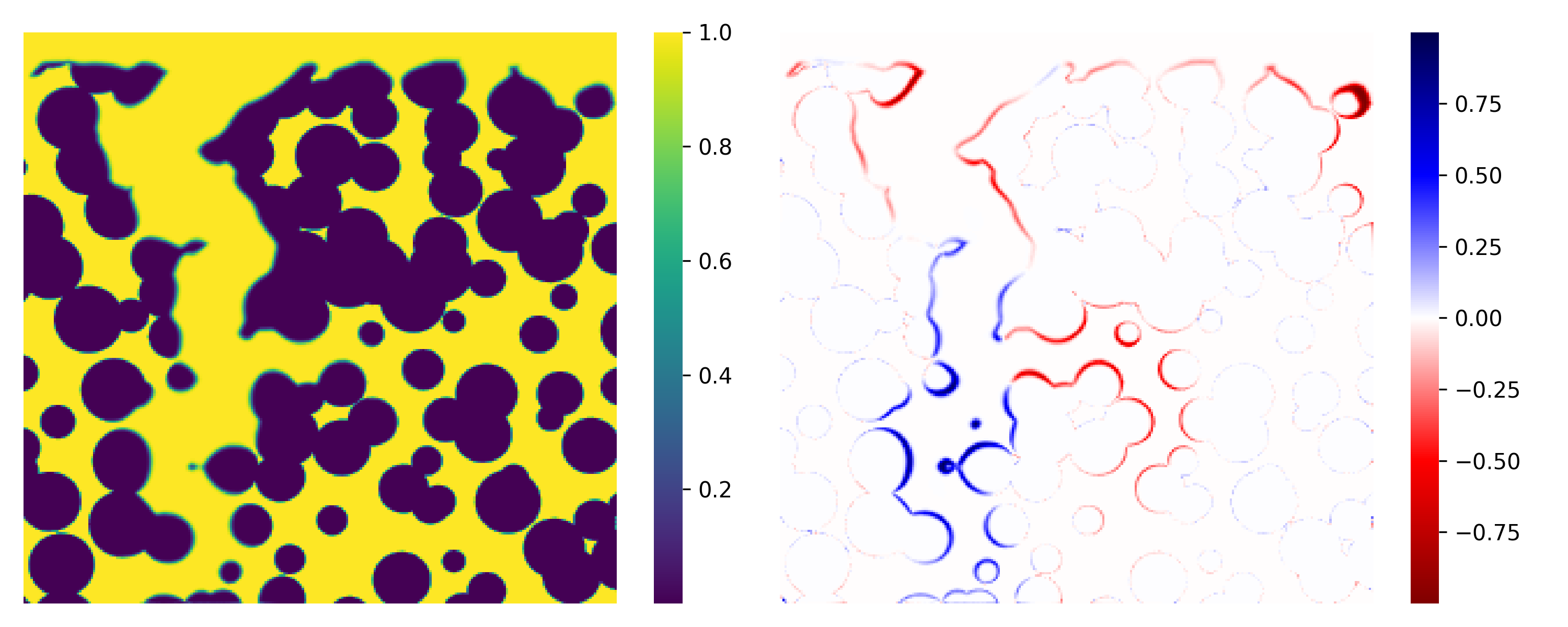}
				\subcaption{U-FNO Level 2\\PCC = 0.988}
				\label{fig:ufno_lv2_eps_1}
			\end{minipage}
			\hfill
			\begin{minipage}[b]{0.325\textwidth}
				\centering
				\includegraphics[width=\textwidth]{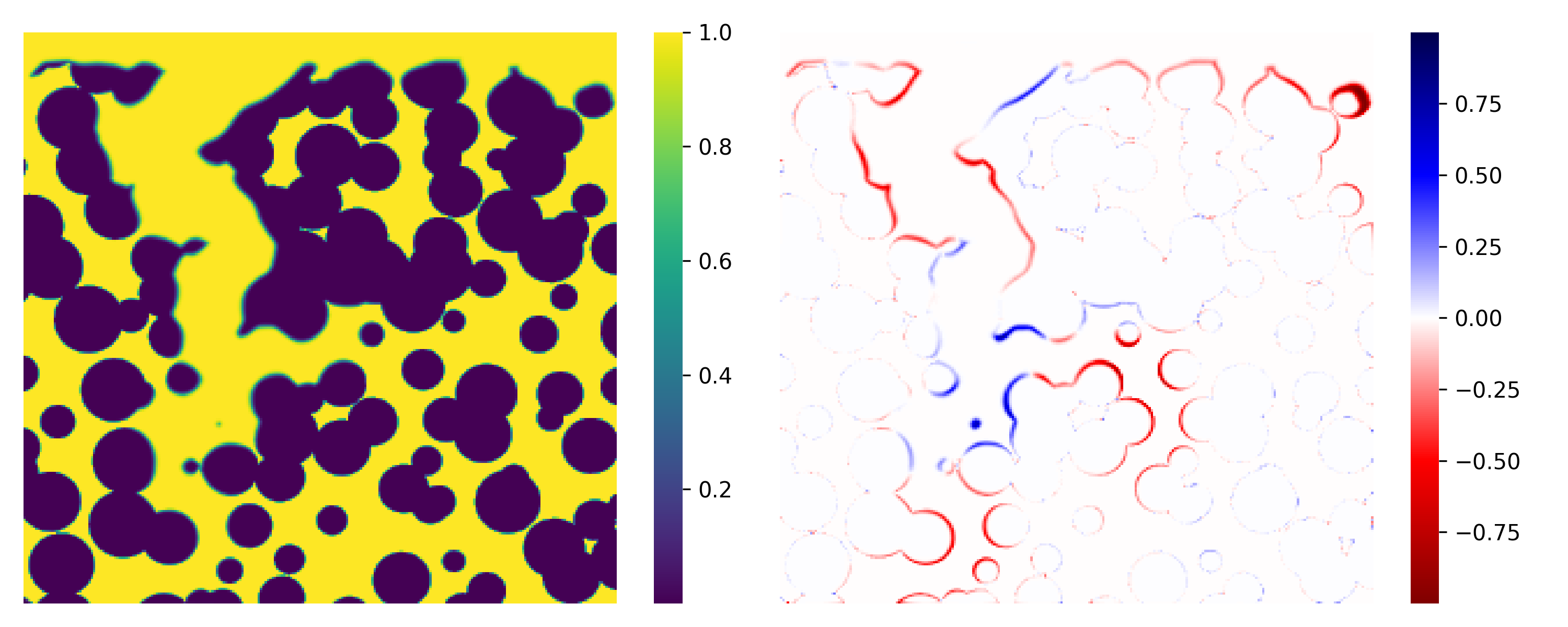}
				\subcaption{TAU Level 2\\\textbf{PCC = 0.992}}
				\label{fig:tau_lv2_eps_1}
			\end{minipage} \\
			\begin{minipage}[b]{0.325\textwidth}
				\centering
				\includegraphics[width=\textwidth]{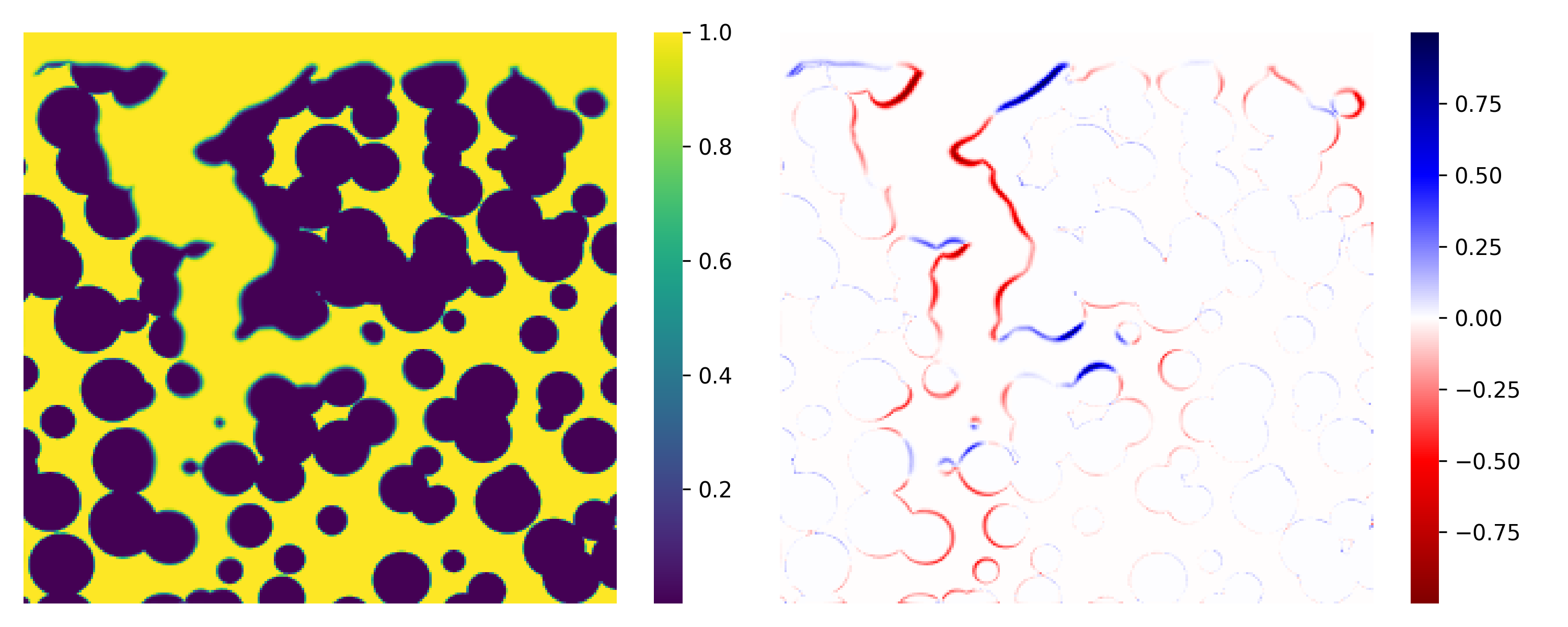}
				\subcaption{ConvLSTM Level 3\\\textbf{PCC = 0.992}}
				\label{fig:convlstm_lv3_eps_1}
			\end{minipage}
			\hfill
			\begin{minipage}[b]{0.325\textwidth}
				\centering
				\includegraphics[width=\textwidth]{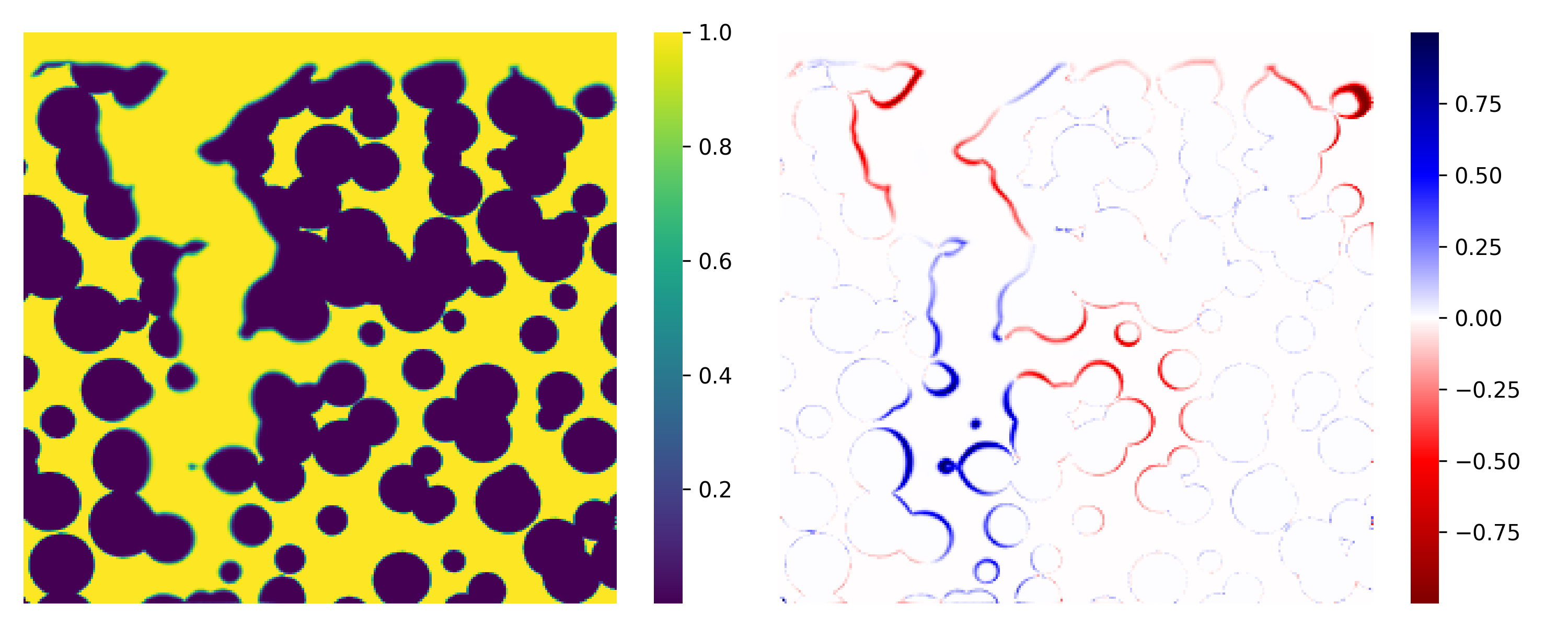}
				\subcaption{U-FNO Level 3\\PCC = 0.987}
				\label{fig:ufno_lv3_eps_1}
			\end{minipage}
			\hfill
			\begin{minipage}[b]{0.325\textwidth}
				\centering
				\includegraphics[width=\textwidth]{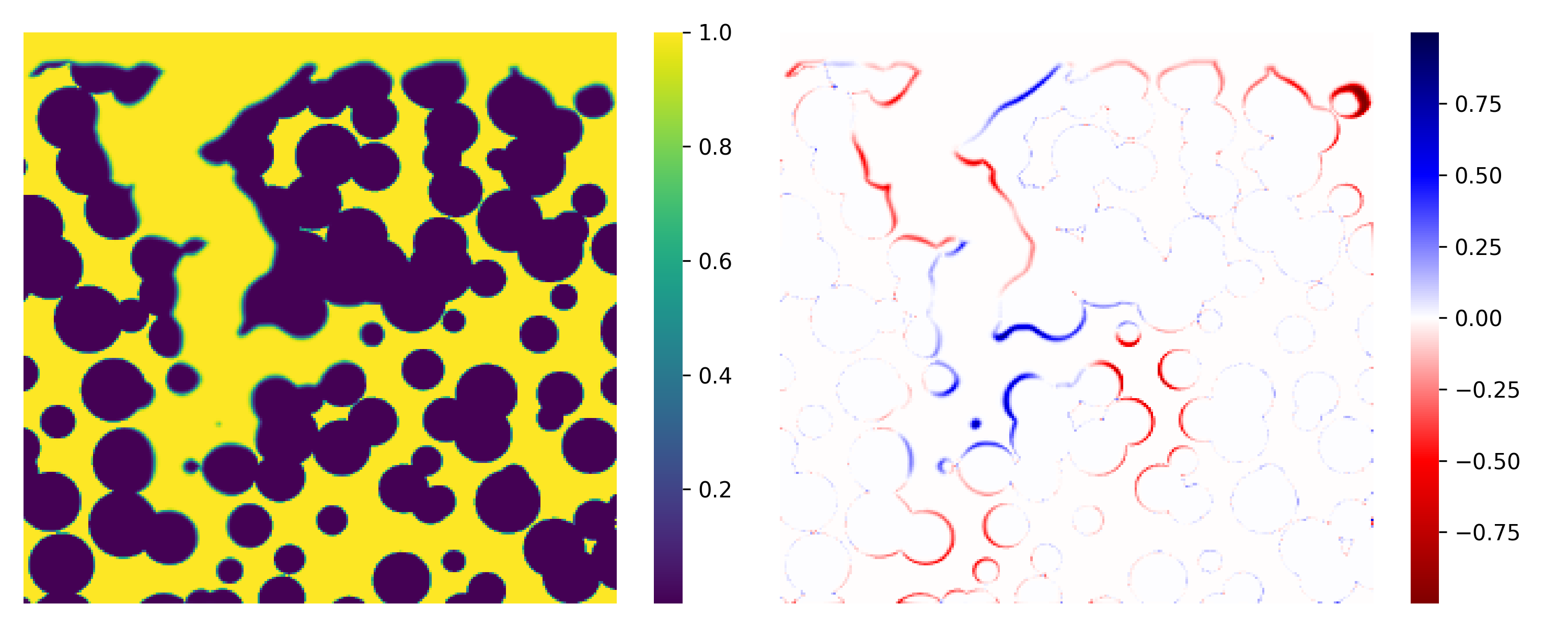}
				\subcaption{TAU Level 3\\\textbf{PCC = 0.992}}
				\label{fig:tau_lv3_eps_1}
			\end{minipage}
			\caption{Predictions of $eps$ at time step 100 for a case from the validation set, along with their respective PCC scores and difference maps to ground truth (Figure~\ref{fig:true_eps_step100}). Best results for each network level are in bold.}
			\label{fig:eps_preds_case1}
		\end{figure}

		Figure~\ref{fig:eps_preds_case1} shows the predictions for $eps$ at time step 100. In general, all networks managed to yield nearly perfect predictions when compared to the ground truth. The PCC scores for ConvLSTM strictly increase over all levels. At Level 0, we can notice a prevalence of overpredictions all over the map, which is mitigated at the subsequent levels. On the other hand, neither U-FNO nor TAU showed significant differences from one level to another, with the latter achieving the best PCC scores at each level, having only tied with ConvLSTM at Level 3.
		
		\subsection{Estimation of Bulk Properties}
		
		To convey a better understanding of the overall dissolution dynamics in our case study, we analyze the evolution of two of the bulk properties used for modelling reactive dissolution at the field-scale: porosity and permeability. For each sample from our validation set, we selected 10 $eps$ maps, starting at time step 5, with an offset of 10 time steps between each pair of consecutive maps,  ending at time step 95. We repeated this process for the $eps$ maps predicted by each of the DL algorithms at the aforementioned time steps. The porosity and permeability values for all maps are then calculated using GeoChemFoam, and the results achieved by each algorithm are compared against the ones obtained from the original data.
		
		
		For all cases discussed in this section, we will show how the errors of porosity and permeability for each algorithm evolve due to the dissolution process in terms of: 1) their respective average values; and 2) the RMSE error versus the values from the ground truth (referred as "Original Data" in the subsequent plots).  
		
		\subsubsection{Porosity Estimation}
		
\begin{figure}[hpbt!]
			\centering
			\begin{minipage}[b]{\textwidth}
				\centering
				\includegraphics[width=\textwidth]{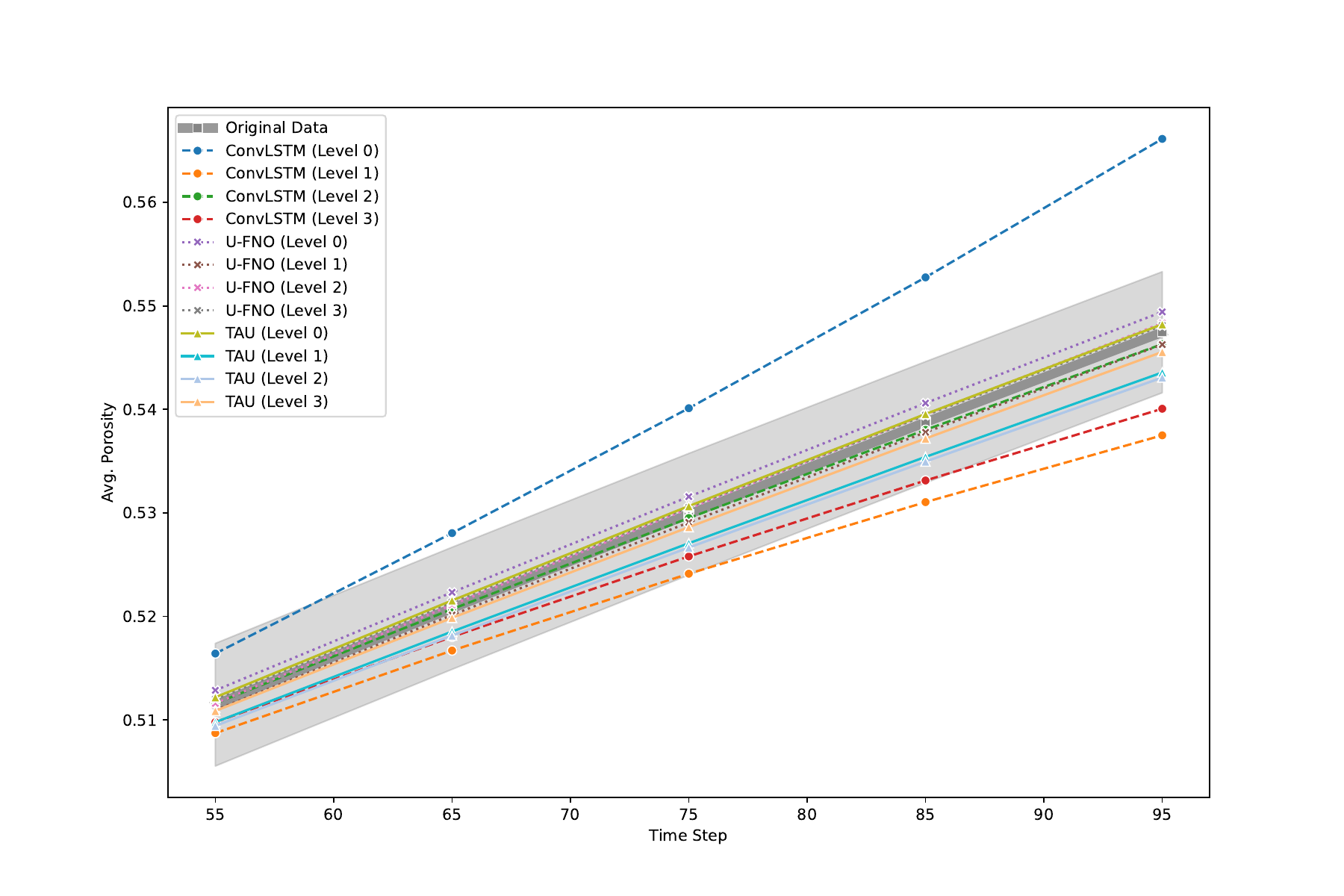}
				\subcaption{Absolute Porosity Evolution (Training Set)}
				\label{fig:avg_poro_train}
			\end{minipage}
			\\
			\begin{minipage}[b]{\textwidth}
				\centering
				\includegraphics[width=\textwidth]{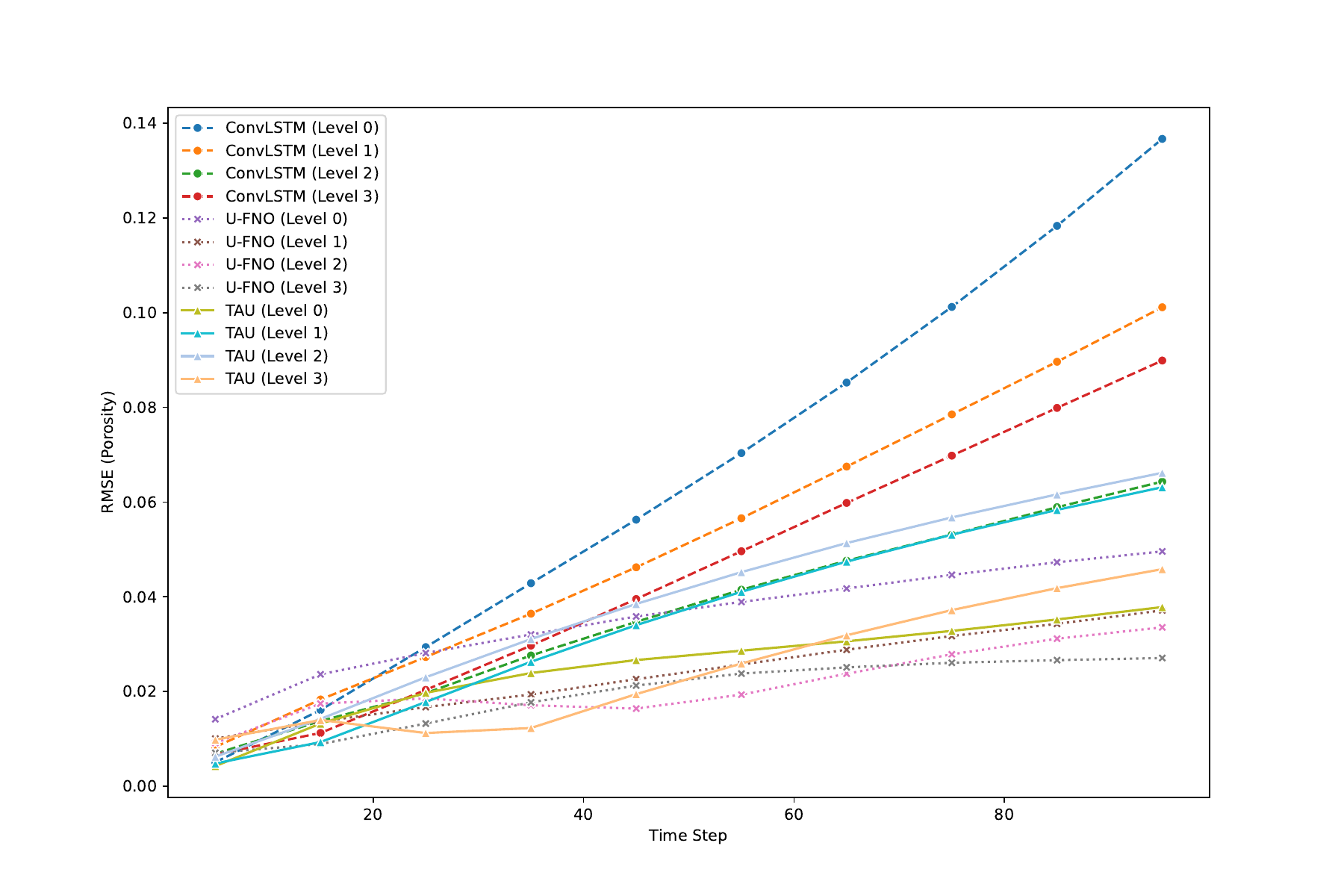}
				\subcaption{Porosity Error Evolution vs. Original Data (Training Set)}
				\label{fig:rmse_poro_train}
			\end{minipage}
			\caption{Porosity error analysis, considering the averages for each time step across all samples from the training set. (\subref{fig:avg_poro_train}) absolute porosity evolutions vs. ground truth (with error margin bounds in grey); (\subref{fig:rmse_poro_train}) RMSE evolutions.}
			\label{fig:poro_errors_train}
		\end{figure}

		Figure~\ref{fig:poro_errors_train} shows the average error in porosity considering all models from the training set. To enhance the overall visualization of the average error curves, we will only display the values from time step 50 onwards. Looking at the average evolution over the sampled time steps (Figure~\ref{fig:avg_poro_train}), we observe that the curves from U-FNO and TAU are the closest from the original porosity values, especially at late time steps, where the error tends to be higher due to the error accumulation problem of the iterative approach. Conversely, the curves from ConvLSTM are farther away from the ground truth, even though it shows some improvement over the network levels. In the RMSE curves (Figure~\ref{fig:rmse_poro_train}), we observe that TAU Level 3 achieved the lowest errors on time steps 25 and 35, but ends up with a higher error than its Level 0 counterpart and all U-FNO variants (except U-FNO Level 0). Both ConvLSTM and U-FNO showed a more consistent evolution over the network levels, whereas the U-FNO Level 3 having lowest errors at the end of the dissolution. 
		
		\begin{figure}[hpbt!]
			\centering
			\begin{minipage}[b]{\textwidth}
				\centering
				\includegraphics[width=\textwidth]{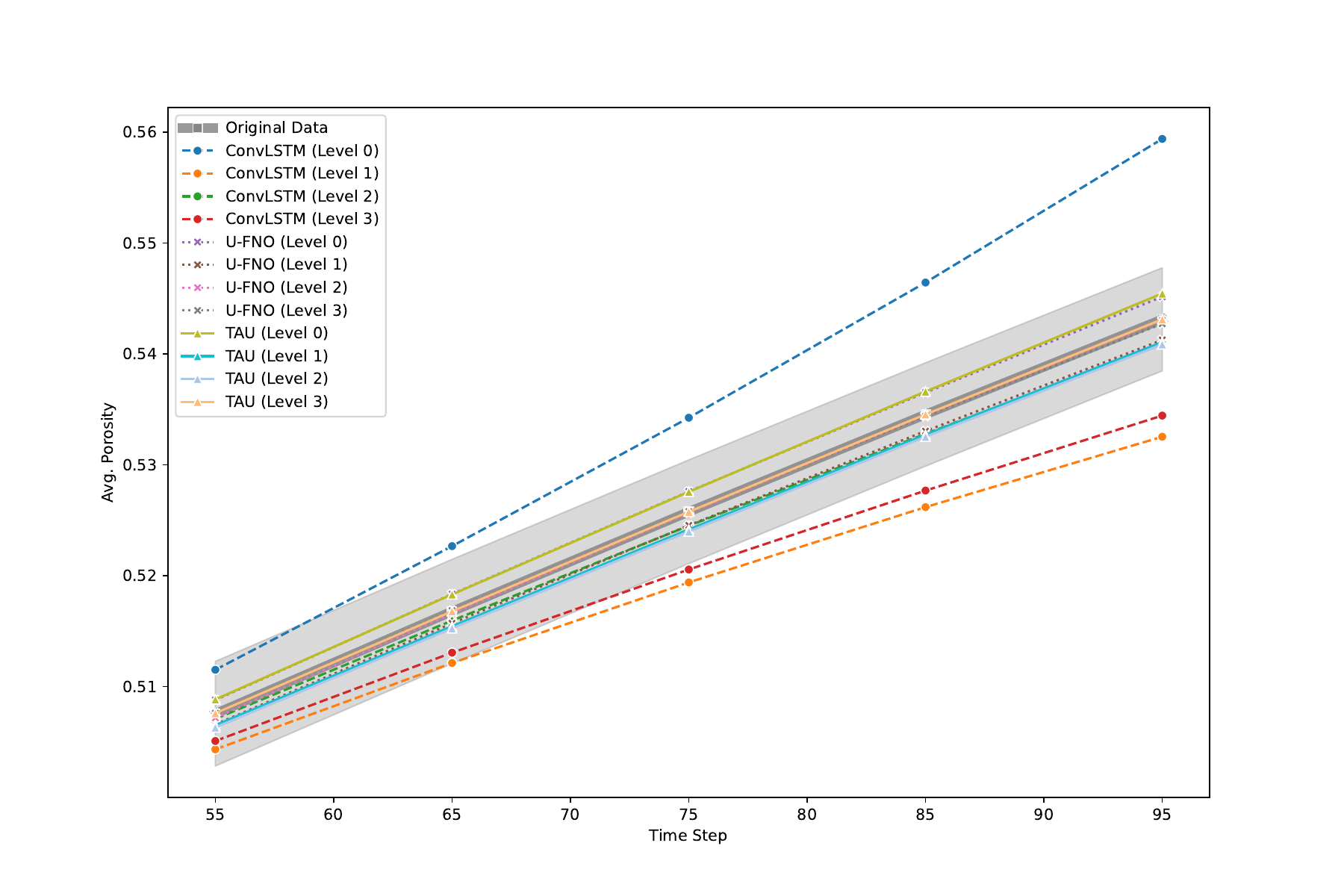}
				\subcaption{Absolute Porosity Evolution (Validation Set)}
				\label{fig:avg_poro_val}
			\end{minipage}
			\\
			\begin{minipage}[b]{\textwidth}
				\centering
				\includegraphics[width=\textwidth]{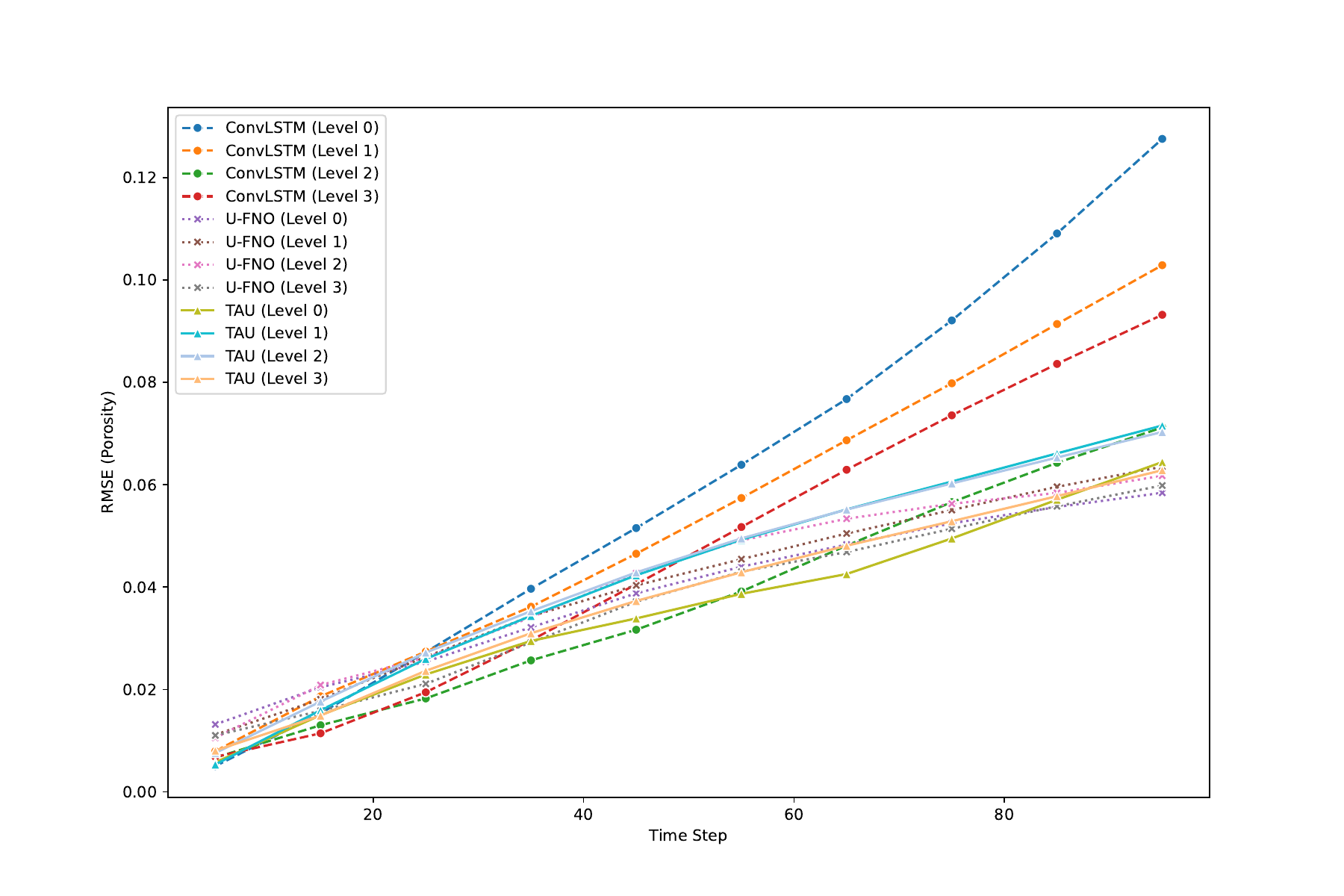}
				\subcaption{Porosity Error Evolution vs. Original Data (Validation Set)}
				\label{fig:rmse_poro_val}
			\end{minipage}
			\caption{Porosity error analysis, considering the averages for each time step across all samples from the training validation set. (\subref{fig:avg_poro_val}) absolute porosity evolutions vs. ground truth (with error margin bounds in grey); (\subref{fig:rmse_poro_val}) RMSE evolutions.}
			\label{fig:poro_errors_val}
		\end{figure}
		
		For the validation set, similar results as the training set are achieved for the average error plots (Figure~\ref{fig:avg_poro_val}). Here, TAU Level 3 was the closest curve to the ground truth at the end of the dissolution, although the curves from its remaining levels and the ones from U-FNO were slightly farther away, but still constantly lying inside the error interval. Concerning the RMSE plots (Figure~\ref{fig:rmse_poro_val}), we can also notice an error reduction on higher network levels for all algorithms, especially after time step 45. This reduction is more noticeable for ConvLSTM, whose Level 2 network achieved significantly lower results than its Level 0 counterpart, at the same time it yielded the lowest errors between time steps 25 and 55. TAU produced very close results among all levels, where Level 2 achieved the lowest error rates at late time steps, but still being slightly worse than ConvLSTM Level 2. Last, U-FNO showed the highest errors at early time steps (5 to 25), achieving the lowest error rates for late time steps (65 to 95).
		
		
		\subsubsection{Permeability Estimation}
		
\begin{figure}[hpbt!]
			\centering
			\begin{minipage}[b]{\textwidth}
				\centering
				\includegraphics[width=\textwidth]{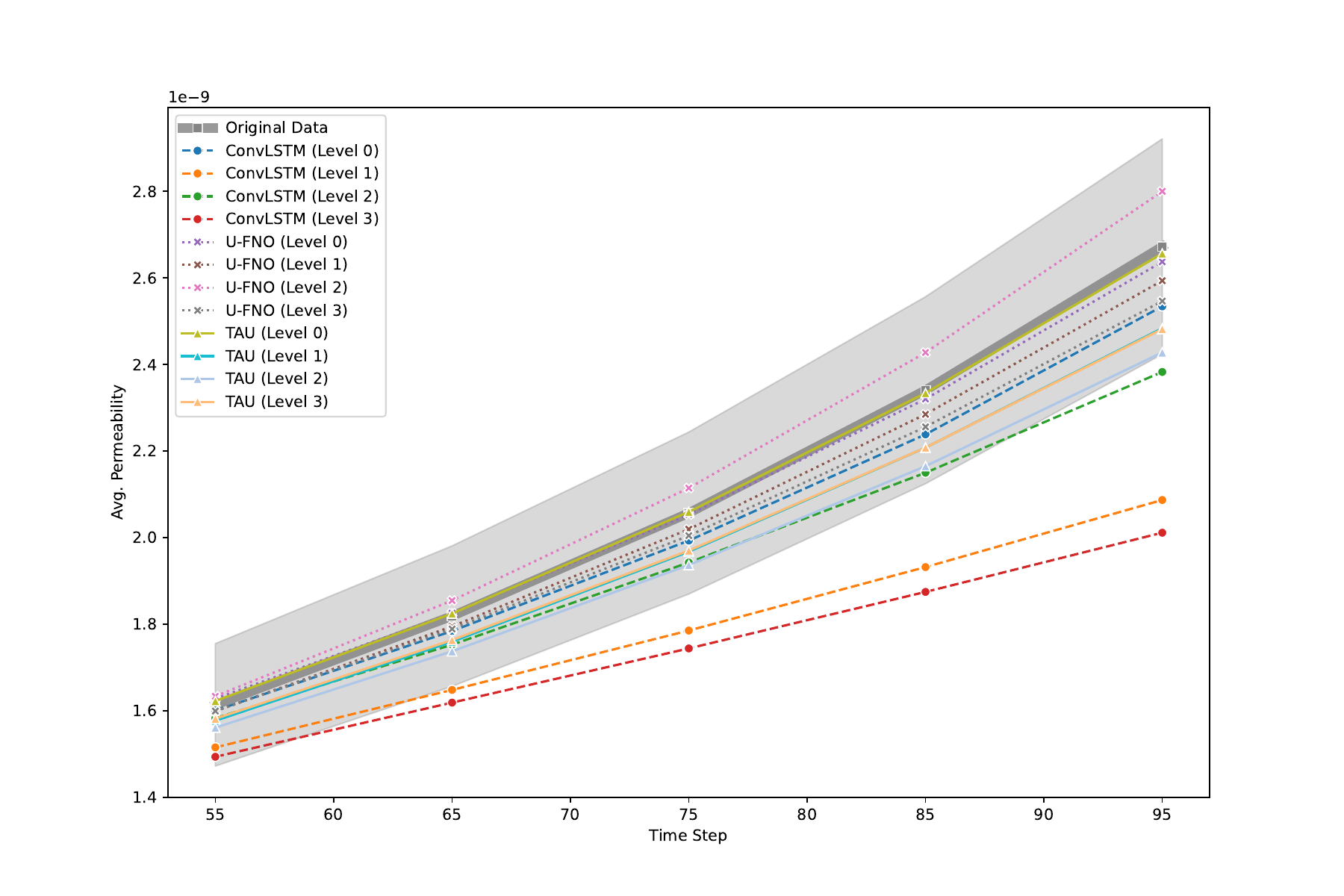}
				\subcaption{Absolute Permeability Evolution (Training Set)}
				\label{fig:avg_perm_train}
			\end{minipage} \\
			\begin{minipage}[b]{\textwidth}
				\centering
				\includegraphics[width=\textwidth]{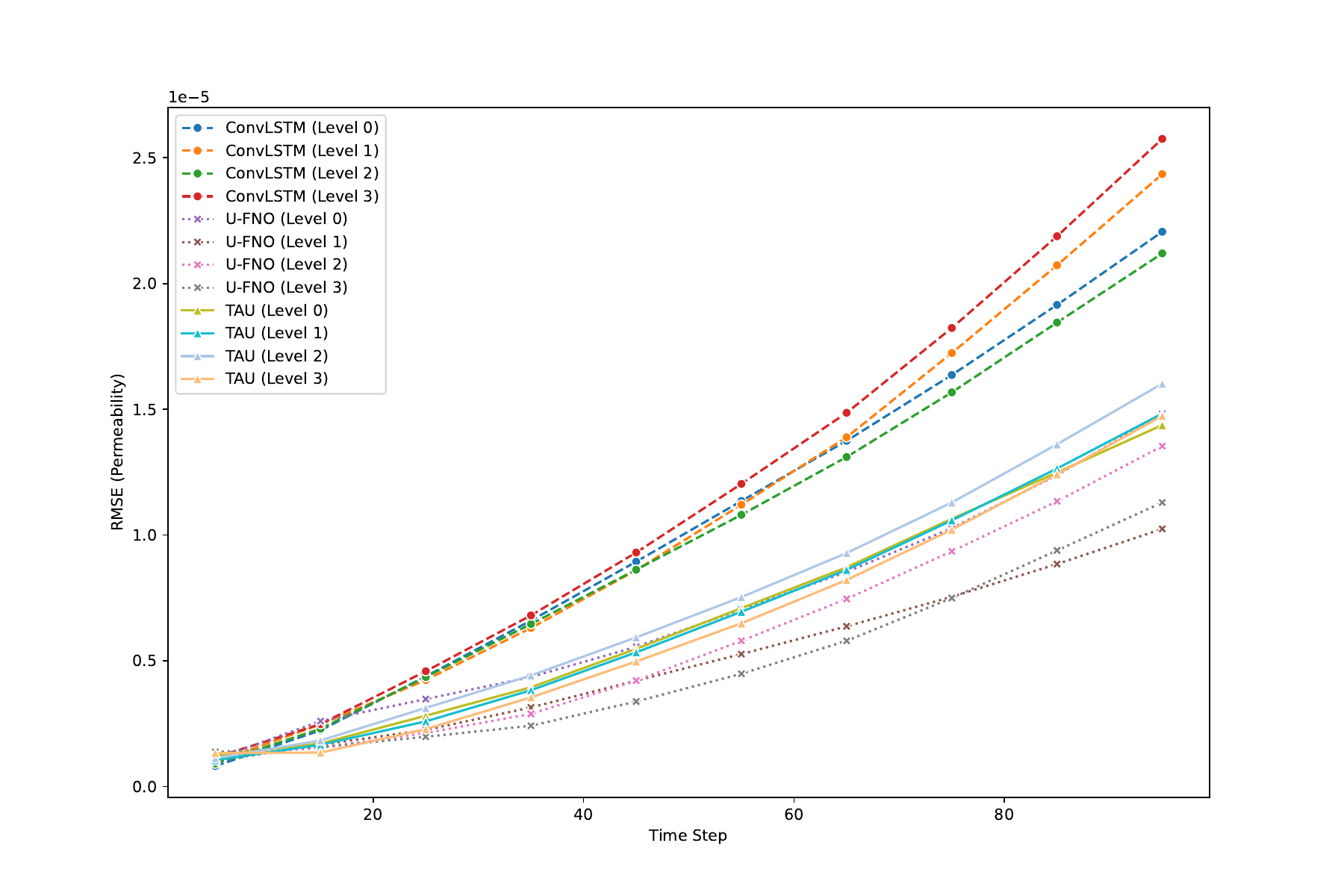}
				\subcaption{Permeability Error Evolution vs. Original Data (Training Set)}
				\label{fig:rmse_perm_train}
			\end{minipage}
			\caption{Permeability error analysis, considering the averages for each time step across all samples from the training set. (\subref{fig:avg_perm_train}) absolute permeability evolutions vs. ground truth (with error margin bounds in grey); (\subref{fig:rmse_perm_train}) RMSE evolutions.}
			\label{fig:perm_errors_train}
		\end{figure}

		Figure~\ref{fig:perm_errors_train} showcases the results for the permeability estimation on the training set. From the curves in both plots, we can observe that the multi-level stacking was not enough to produce a consistent evolution over the levels for all algorithms. This is corroborated by the fact that the closest curves to the ground truth (Figure~\ref{fig:avg_perm_train}) were obtained from TAU Level 0 and U-FNO Level 0. Moreover, when analyzing the RMSE curves for each algorithm (Figure~\ref{fig:rmse_perm_train}), ConvLSTM Level 3 yielded the worst results during all the dissolution steps. Regarding the TAU curves, the Level 3 network achieved the lowest error rates until the last time step, where the Level 0 network was slightly better. For the U-FNO, the Level 3 network was the best among all networks between time steps 25 and 65, being later surpassed by its Level 1 counterpart.
		
		\begin{figure}[hpbt!]
			\centering
			\begin{minipage}[b]{\textwidth}
				\centering
				\includegraphics[width=\textwidth]{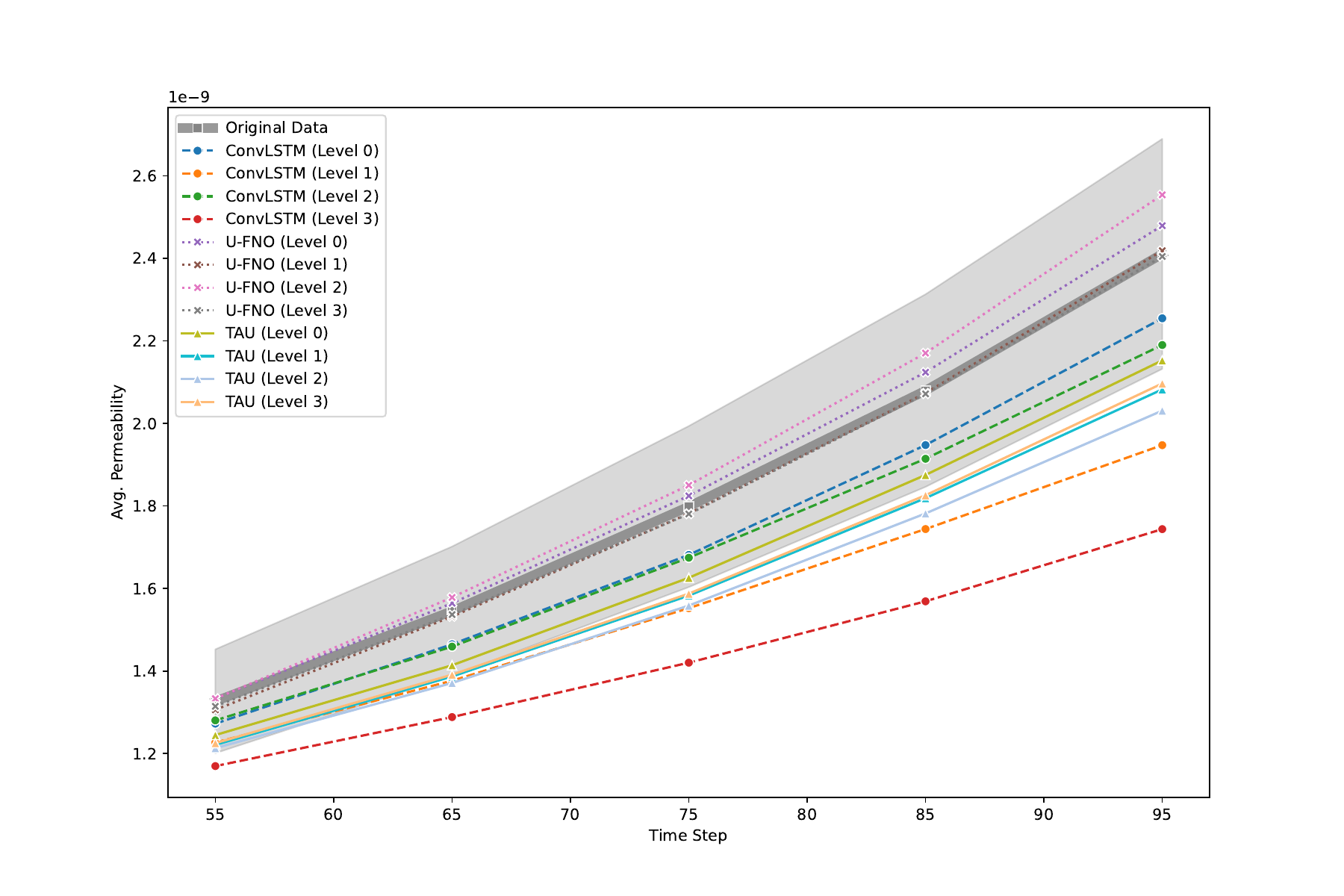}
				\subcaption{Average Permeability Evolution (Validation Set)}
				\label{fig:avg_perm_val}
			\end{minipage} \\
			\begin{minipage}[b]{\textwidth}
				\centering
				\includegraphics[width=\textwidth]{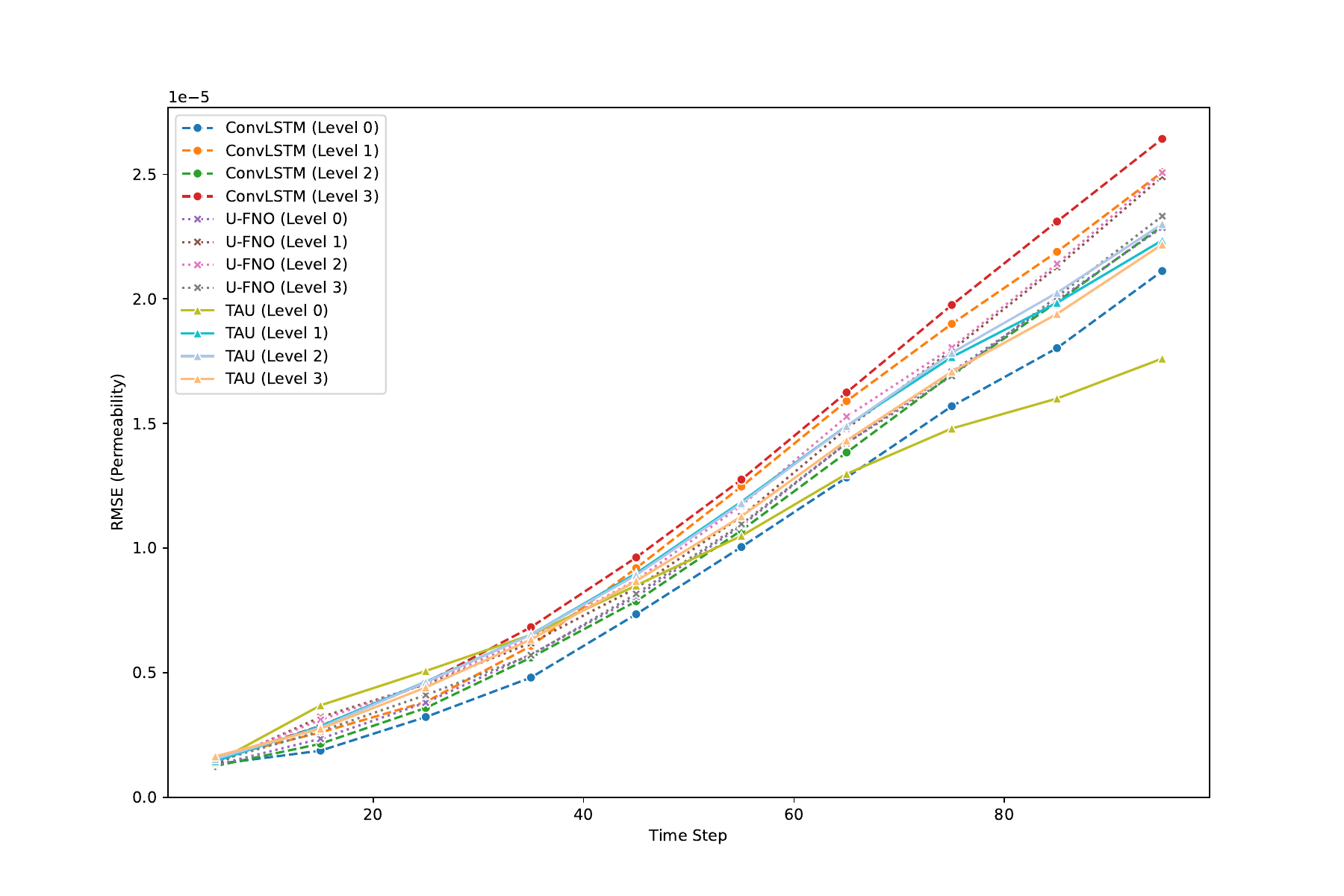}
				\subcaption{Permeability Error Evolution vs. Original Data (Validation Set)}
				\label{fig:rmse_perm_val}
			\end{minipage}
			\caption{Permeability error analysis, considering the averages for each time step across all samples from the validation set. (\subref{fig:avg_perm_val}) absolute permeability evolutions vs. ground truth (with error margin bounds in grey); (\subref{fig:rmse_perm_val}) RMSE evolutions.}
			\label{fig:perm_errors_val}
		\end{figure}
		
		
		
		With respect to the estimation on the validation set, the evolution curves from Figure~\ref{fig:avg_perm_val} show that the U-FNO curves were the closest ones to the ground truth. Unlike the training set, the TAU curves were farther away from the ground truth, where the Level 0 curve was the only one to remain inside the error margin bounds during all the dissolution. About ConvLSTM, the best results were achieved by its Level 0 network.
		
		Considering the RMSE curves from Figure~\ref{fig:rmse_perm_val}, the lowest errors were achieved by \-ConvLSTM Level 0 (time steps 15 to 45) and TAU Level 0 (time steps 55 to 95). Furthermore, ConvLSTM Level 3 produced the highest errors between time steps 45 to 95. For the same interval, U-FNO Level 3 achieved the lowest results among all its levels, although still being slightly worse than TAU Level 0.
		
		
		\section{Conclusions}
		\label{sec:conc}
		
		This paper presented a data-driven method that leverages deep learning methods to predict the evolution in time of reactive dissolution in porous media. An iterative stacked MIMO approach was adopted to produce the future states of a dissolution process, starting with an initial set of "perfect" inputs, generating an output which is then used as input to predict the subsequent states, and so on. To mitigate the overall errors of the predictions, a multi-level stacking approach was proposed, where each level is trained to correct the errors produced by the previous level network. Three different algorithms (ConvLSTM, U-FNO and TAU) were tested in a dataset comprised of 32 numerical simulation models.
		
		Although error accumulation in recursive strategies for prediction of time-series data is still an open problem, all algorithms showed high correlation scores, especially regarding the predictions of $C$ and $eps$, even at late time steps. Moreover, the multi-stacking approach was successful at improving the results from the base model (Level 0) of each algorithm for the majority of the cases. On the other hand, TAU did not benefit so much from this pipeline, which emphasizes the need for further investigation on the weight of the regularization term of its loss function. Even so, its Level 0 network was capable of achieving higher correlations than the other algorithms (and their corrections) for all predicted properties (except $U_x$), at the same time it achieved faster training and forward times.
		
		Despite the high correlation scores for $eps$ prediction, there is still some improvement possible in estimating  bulk properties from the predicted $eps$ maps, which would bridge the gap between pore-scale interactions and macroscopic flow and transport behaviors, and provide a broader understanding of such phenomena in a porous medium. This is particularly true when assessing the error rates for porosity and permeability estimation, which were expected to decrease over each network level, following similar patterns to the $eps$ correlation plots. However, as none of the networks was calibrated to be aware of the overall bulk properties at a given time step, the evolution of those error rates ended up showing an uncorrelated pattern to the $eps$ predictions. Nevertheless, all algorithms managed to produce low-magnitude error rates for porosity and permeability, and similar evolutions to their respective ground truths, considering an "average pore geometry" across all samples from our dataset. Hence, our method has a potential to replace traditional numerical solvers, especially when also taking into account the reported speedup and lowered computational expense.
		
		Future directions for this work include: 1) analysis of gradient accumulation to improve iterative predictions at late time steps; 2) in-depth study of bulk property estimation; 3) application of the proposed method in larger-scale domains and in 3-dimensional simulations.

				\acknowledgments
				This work is funded by the Engineering and Physical Sciences Research Council's ECO-AI Project grant (reference number EP/Y006143/1), with additional financial support from the PETRONAS Centre of Excellence in Subsurface Engineering and Energy Transition (PACESET).
				
				\section*{Author Contributions}
				
				The contributions of each of the authors of this manuscript are listed as follows:
				
				\begin{itemize}
					\item Marcos Cirne: formal analysis, investigation, methodology, software, validation, visualization, writing (original draft);
					\item Hannah P. Menke: conceptualization, data curation, formal analysis, methodology, supervision, validation, writing (review \& editing);
					\item Alhasan Abdellatif: software, validation;
					\item Julien Maes: conceptualization, resources, validation, writing (review \& editing);
					\item Florian Doster: conceptualization, supervision, validation, writing (review \& editing);
					\item Ahmed H. Elsheikh: conceptualization, funding acquisition, methodology, project administration, resources, supervision, validation, writing (review \& editing);
				\end{itemize}
				
				\section*{Data Availability Statement}
				
				The source code used to reproduce all results for our iterative stacked method, including pre-trained models for all ML algorithms described in this work, can be found at \url{https://github.com/ai4netzero/ReactiveDissolution}. The supporting dataset for reactive dissolution is publicly available at \url{https://zenodo.org/records/14974428} under the Creative Commons Attribution International 4.0 license~\cite{reactdataset2025}.
				
				\section*{Software Availability Statement}
				
				The experiments on porosity and permeability estimation were run on version 5.1 of GeoChemFoam, available at \url{https://github.com/GeoChemFoam} under the GNU General Public License (GPL-3.0)~\cite{maes2022improved}.
				
				\section*{Conflicts of Interest}
				
				We have no conflicts of interest to disclose in this manuscript.
				
				%
				%
				
				
				\bibliography{references}
				
				%
				%
				%
				%
				%

				%
				%
				%
				%
				\appendix
				\section{Mean Squared Error (MSE) Scores for Iterative Prediction}
				\label{app:mse}
				
\begin{figure}[h!]
					\centering
					\includegraphics[width=0.89\linewidth]{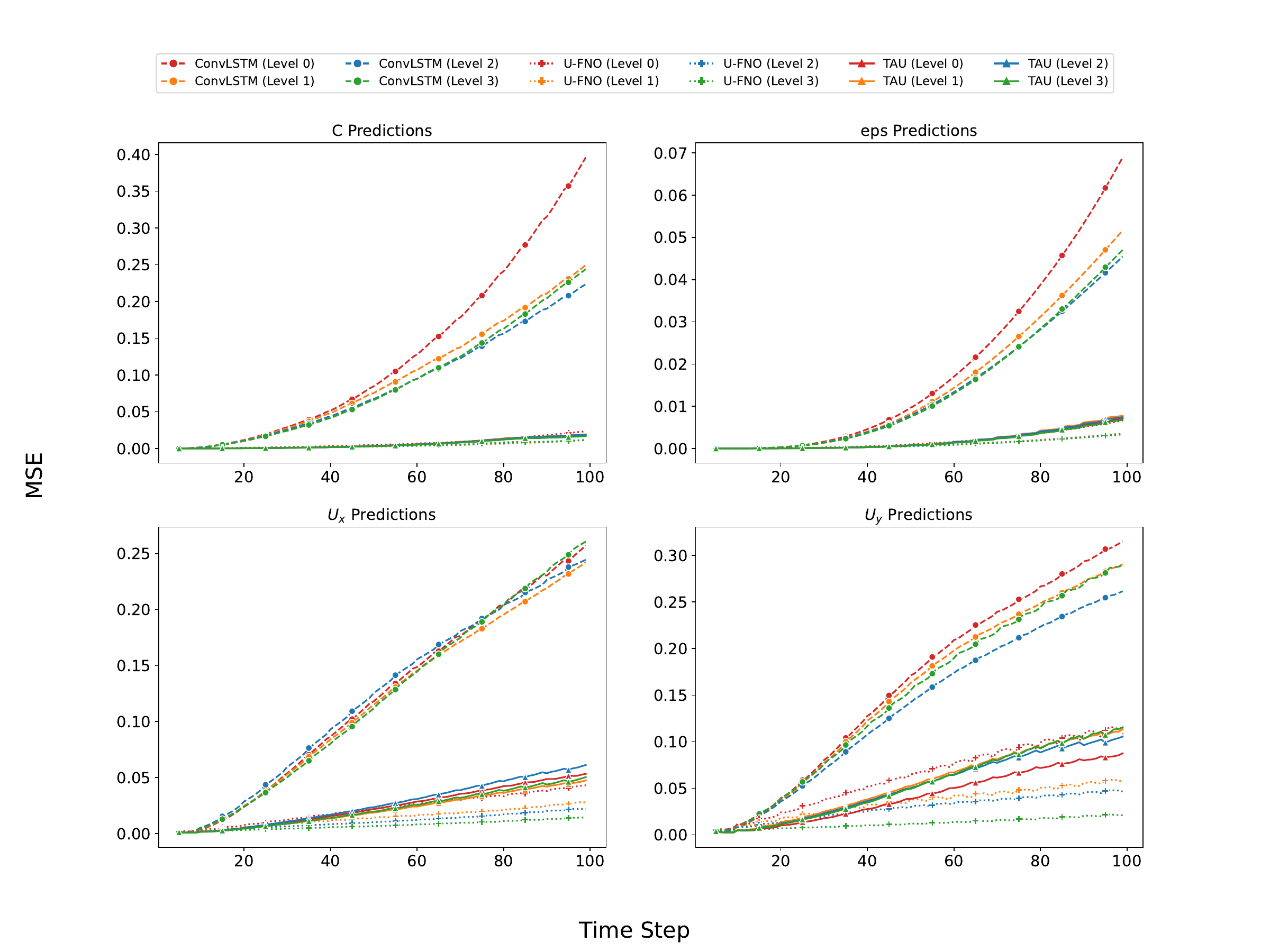}
					\caption{Average MSE scores of all samples from the training set for the iterative predictions produced by each algorithm.}
					\label{fig:mse_train}
				\end{figure}

\begin{figure}[h!]
					\centering
					\includegraphics[width=0.89\linewidth]{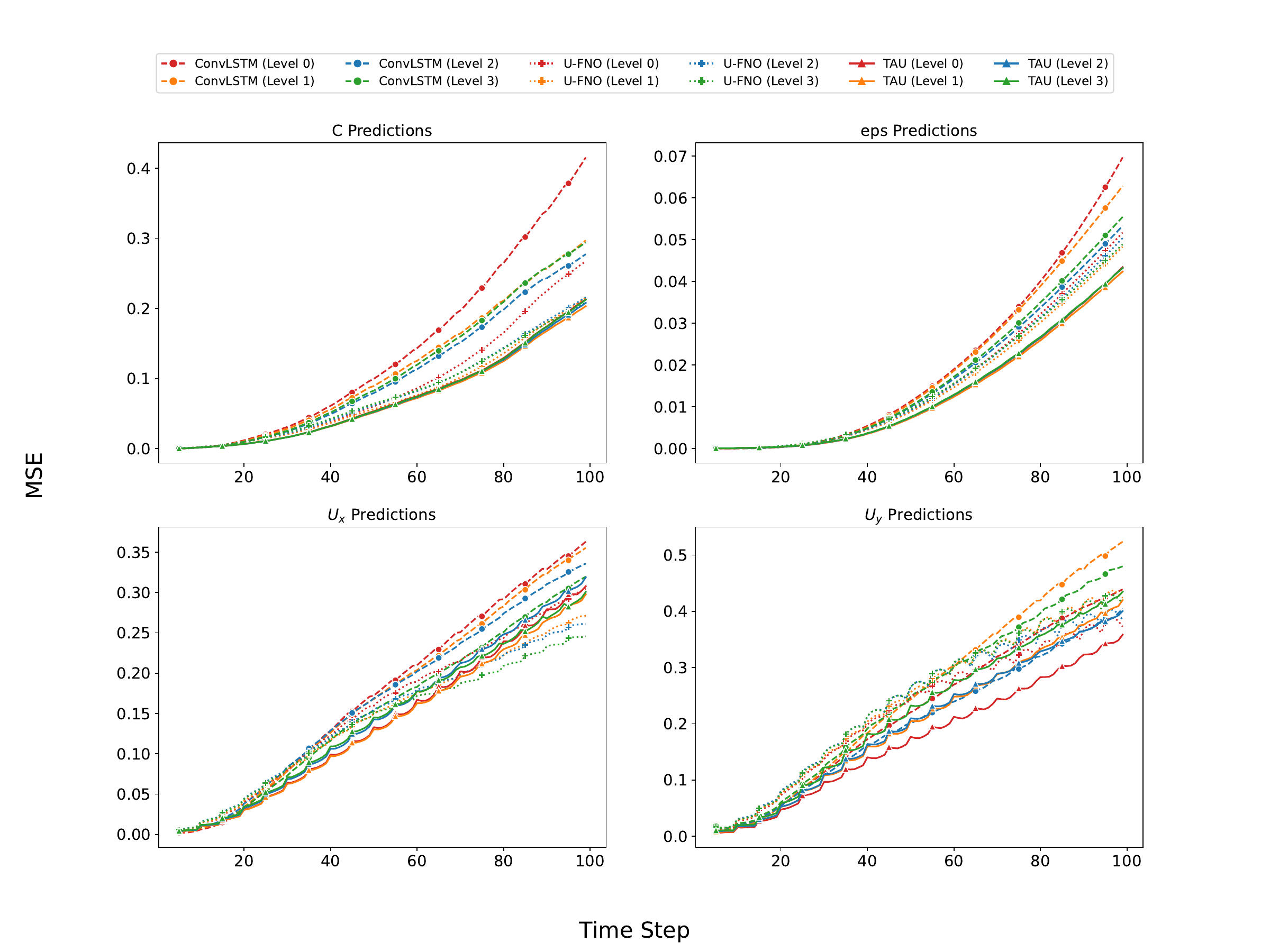}
					\caption{Average MSE scores of all samples from the validation set for the iterative predictions produced by each algorithm.}
					\label{fig:mse_val}
				\end{figure}

				To quantify the error magnitudes of the iterative predictions, we also conducted an analysis of the evolution of MSE scores. Figures~\ref{fig:mse_train} and~\ref{fig:mse_val} show the MSE scores of the iterative predictions on the training and validation sets, respectively. Compared to the results discussed in Section~\ref{sec:rec_pred}, the ranking of all methods on both scenarios remained the same as the ones produced by the PCC metric. These results indicate that TAU not only has a higher linear relationship to the ground truth, but also yields the smallest errors on its predictions among the tested algorithms.
				
			\end{document}